\documentclass{article}

\usepackage{microtype}
\usepackage{graphicx}
\usepackage{subcaption}
\usepackage{booktabs} 
\usepackage{xspace}
\usepackage{multicol}
\usepackage{multirow}
\usepackage{esvect}
\usepackage{stmaryrd}
\usepackage{enumitem}

\usepackage{hyperref}

\newcommand{\proposal}{\textit{SleepMaMi}\xspace}
\newcommand{\edit}[1]{\textcolor{black}{#1}}

\usepackage[accepted]{icml2026}

\usepackage{amsmath}
\usepackage{amssymb}
\usepackage{mathtools}
\usepackage{amsthm}

\usepackage[capitalize,noabbrev]{cleveref}

\theoremstyle{plain}

\theoremstyle{definition}

\theoremstyle{remark}

\usepackage[textsize=tiny]{todonotes}

\icmltitlerunning{SleepMaMi: A Universal Sleep Foundation Model for Integrating Macro- and Micro-structures}

\begin{document}

\twocolumn[
  \icmltitle{SleepMaMi: A Universal Sleep Foundation Model for Integrating Macro- and Micro-structures}



  \icmlsetsymbol{equal}{*}

  \begin{icmlauthorlist}
    \icmlauthor{Keondo Park}{a}
    \icmlauthor{Younghoon Na}{b,c}
    \icmlauthor{You Rim Choi}{a}
    \icmlauthor{Hyunwoo Ryu}{a}
    
    \icmlauthor{Hyun-Woo Shin}{b,c,d}
    \icmlauthor{Hyung-Sin Kim}{a}
  \end{icmlauthorlist}

  \icmlaffiliation{a}{Graduate School of Data Science, Seoul National University, Seoul, South Korea}
  \icmlaffiliation{b}{Department of Biomedical Sciences, Seoul National University College of Medicine, Seoul, Republic of Korea}
  \icmlaffiliation{c}{Obstructive Upper Airway Research (OUaR) Laboratory, Department of Pharmacology, Seoul National University College of Medicine, Seoul, Republic of Korea}
  \icmlaffiliation{d}{Department of Otorhinolaryngology-Head and Neck Surgery, Seoul National University Hospital, Seoul, Republic of Korea}

  \icmlcorrespondingauthor{Hyun-Woo Shin}{charlie@snu.ac.kr}
  \icmlcorrespondingauthor{Hyung-Sin Kim}{hyungkim@snu.ac.kr}

  \icmlkeywords{Sleep Foundation Model, ICML}

  \vskip 0.3in
]



\printAffiliationsAndNotice{}  

\begin{abstract}

While the shift toward unified foundation models has revolutionized many deep learning domains, sleep medicine remains largely restricted to task-specific models that focus on localized micro-structure features. These approaches often neglect the rich, multi-modal context of Polysomnography (PSG) and fail to capture the global macro-structure of a full night's sleep. To address this, we introduce \proposal, a Sleep Foundation Model engineered to master both hour-long sleep architectures and fine-grained signal morphologies. Our framework utilizes a hierarchical dual-encoder design: a Macro-Encoder to model full-night temporal dependencies and a Micro-Encoder to capture short-term characteristics from biosignals. Macro-Encoder is trained via \textit{Demographic-Guided Contrastive Learning}, which aligns overnight sleep patterns with objective subject metadata, such as age, sex, and BMI to refine global representations. Micro-Encoder is optimized via a hybrid Masked Autoencoder (MAE) and multi-modal contrastive objective.  Pre-trained on a massive corpus of $>$20,000 PSG recordings (158K hours), \proposal outperforms \edit{or matches state-of-the-art} foundation models across a diverse suite of downstream tasks, demonstrating superior generalizability and label-efficient adaptation for clinical sleep analysis. 
\end{abstract}

\section{Introduction}

Sleep occupies approximately one-third of the human lifespan and is a fundamental pillar of systemic health. 
Given its critical role, extensive research has been dedicated to decoding the complex physiological processes that govern sleep and addressing the myriad sleep disorders affecting hundreds of millions of population worldwide~\cite{benjafield2019estimation}. 
The clinical gold standard for these investigations is Polysomnography (PSG), a multi-modal diagnostic tool that captures a diverse array of biosignals, including electroencephalogram (EEG), electrooculogram (EOG), electrocardiogram (ECG), electromyogram (EMG), and respiratory signals.

Comprehensive sleep analysis necessitates a dual-perspective approach. 
It requires the examination of sleep micro-structure — high-frequency physiological features observable within few seconds — and sleep macro-structure — the global organization of sleep stages and cycles across a full-night recording. 

However, analyzing PSG is not a standard long-sequence modeling problem. It requires integrating heterogeneous modalities with varying sampling rates while capturing both transient micro-events and hour-scale macro-architecture within a single representation. ~\cite{phan2019seqsleepnet} 
\begin{figure*}[t]
    \vspace{-2ex}
    \centering
    \includegraphics[width=0.95\textwidth, bb=0 0 1914 863]{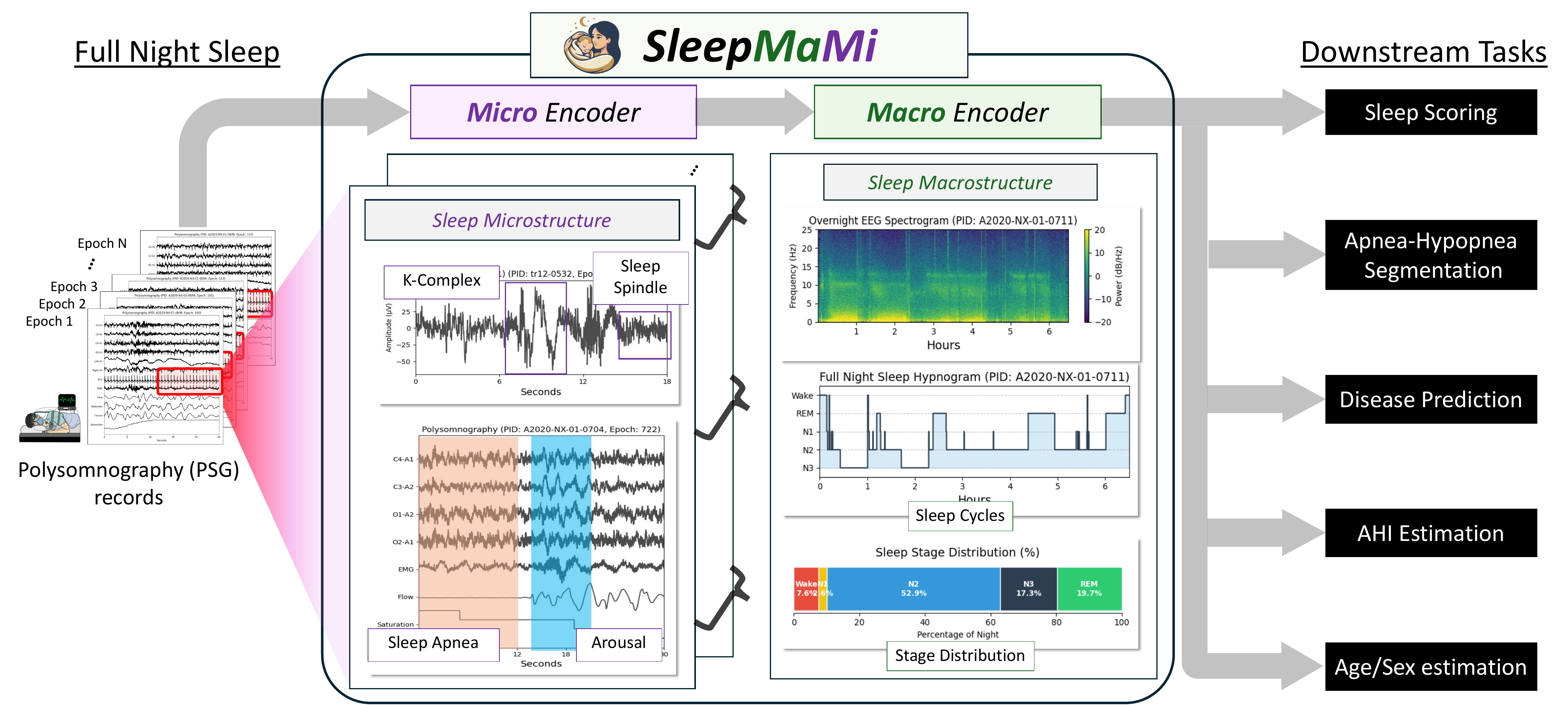}
    \caption{\textbf{Overview of \proposal.} Full-night PSG recordings are processed through a hierarchical dual-encoder architecture. The Micro-Encoder captures short-term physiological patterns such as K-complexes, sleep spindles, and respiratory events, while the Macro-Encoder models global sleep architecture including sleep cycles and stage distributions across the entire recording. This design supports diverse downstream tasks ranging from fine-grained event segmentation to subject-level clinical outcomes.}
    \label{fig:overview}
    \vspace{-3ex}
\end{figure*}

The labor-intensive nature of manual PSG scoring has catalyzed the development of deep learning models for automated analysis. 
However, existing literature remains largely fragmented. 
Most current approaches are task-specific  (e.g., sleep staging) and depend on supervised approaches~\cite{park2025distillsleep, lee2025explainable, retamales2024towards}. 
While these models achieve impressive performance within their narrow domains, they lack the generalized representations necessary for cross-task transfer and depend on expert-scored labels~\cite{perslev2021u}, lacking scalability and exposing the models to annotation noise~\cite{danker2004interrater, danker2009interrater, guillot2020dreem}.

 Crucially, clinical research has established that sleep macro-architecture serves as a vital biomarker for long-term health outcomes and disease prognosis ~\cite{mander2017sleep, blackwell2011associations}. Despite this clinical evidence, a significant gap remains in modeling macro-structural dynamics, which limits the applicability of these models for subject-level clinical diagnosis and long-term health prediction.

To bridge these gaps, we propose \textbf{\proposal}, a unified Sleep Foundation Model designed to master both micro- and macro-structural sleep representations (Figure~\ref{fig:overview}). 
Pre-trained on 20,964 PSG recordings (158,028 hours), \proposal shows strong generalizability and label-efficient adaptation to downstream tasks.
\proposal adopts a hierarchical architecture comprising two specialized components: the \textbf{Micro-Encoder} utilizes a shared-private transformer architecture to capture the idiosyncratic patterns of individual modalities alongside their cross-modal correlations. 
It is optimized via a hybrid strategy of masked autoencoding (MAE)~\cite{he2022masked} and contrastive learning (CL)~\cite{oord2018representation} to learn robust, fine-grained latent features. 
To capture global context, the \textbf{Macro-Encoder} processes full-night sequences using a \textit{Demographic-Guided Contrastive Learning} objective. 
By aligning sequence-level embeddings with objective demographic metadata (e.g., age, sex and BMI), the model learns to represent the normative and pathological trajectories of sleep macrostructure.

Our primary contributions are summarized as follows:
\begin{itemize}[leftmargin=1.0em, itemsep=0pt, topsep=0pt, parsep=0pt, partopsep=0pt]
   \item We introduce \proposal, the first foundation model that explicitly encodes both micro- and macro-structural sleep features, enabling a range of applications from second-level segmentation to subject-level disease prediction.
    \item  We propose a novel training paradigm for biosignals that combines MAE-based reconstruction with multi-modal contrastive alignment, ensuring both signal-specific fidelity and cross-modal consistency.
    \item We introduce \textit{Demographic-Guided Contrastive Learning}, an objective, metadata-driven pre-training strategy that optimizes long-sequence modeling without the need for subjective manual annotations, effectively capturing global sleep architecture.
    \item We demonstrate that \proposal \edit{outperforms or matches} existing foundation models across a diverse suite of benchmarks, proving its robustness in both local and global temporal tasks.
\end{itemize}

\edit{Our source code and checkpoint are available at \url{https://github.com/keondopark/SleepMaMi}}

\section{Related Work}

\begin{figure*}[t]
    \centering
    \vspace{-2ex}
    \includegraphics[width=0.9\textwidth, bb= 0 0 953 535]{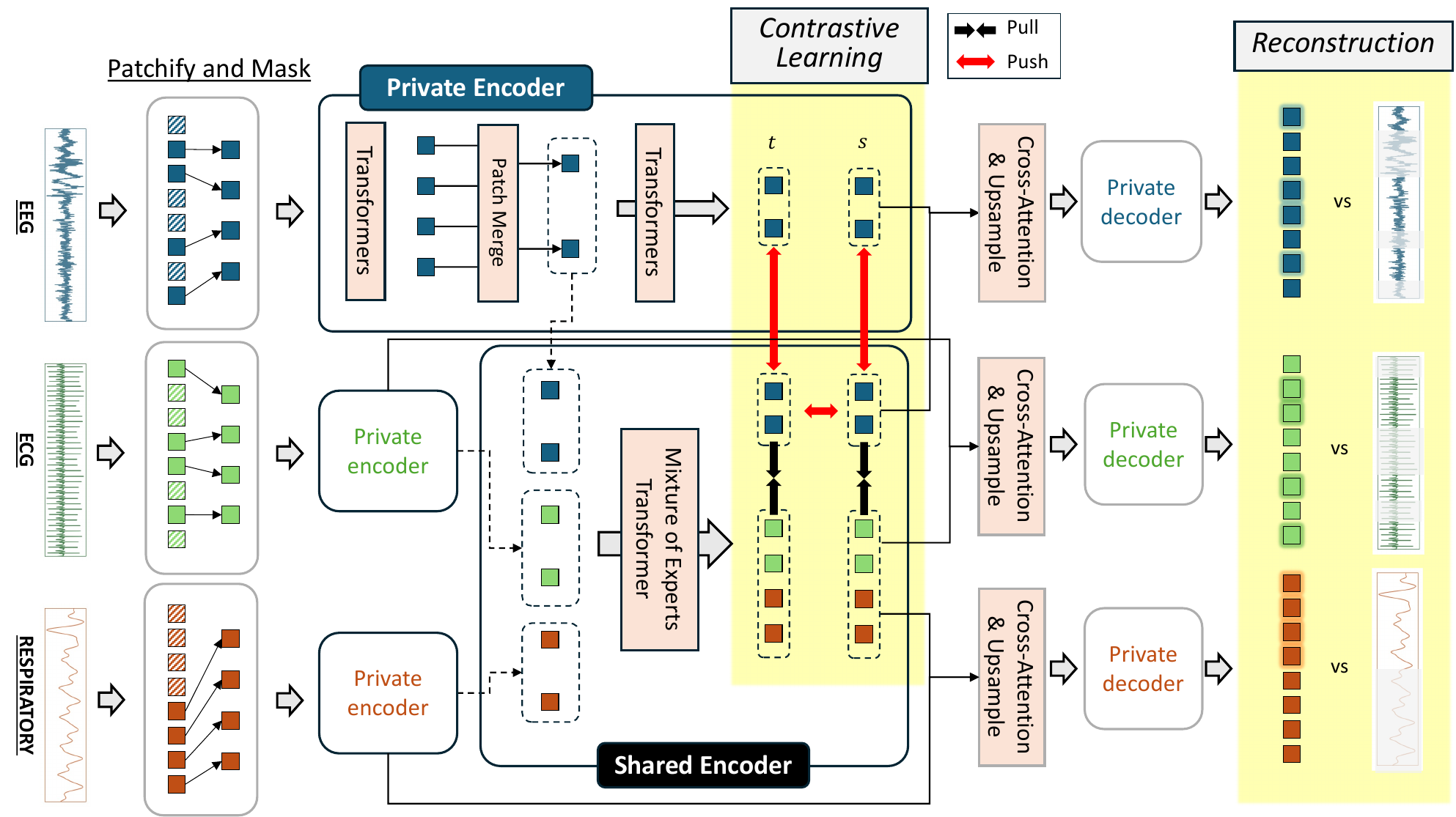}
    \caption{\textbf{Micro-Encoder design and pretraining method.} 
    The Micro-Encoder adopts a private--shared encoder architecture, incorporating patch merging to improve computational efficiency in the shared encoder. The model is trained with a hybrid objective that combines masked autoencoding (reconstruction) and multi-modal contrastive learning to capture sleep micro-structure. For clarity and space constraints, only three representative modalities are shown.}
    \label{fig:micro_encoder}
    \vspace{-3ex}
\end{figure*}

\subsection{Neural Networks for PSG Analysis}
The vast amount of analysis required to understand PSG has drawn a great amount of attention towards using neural networks for the analysis. 
One of the most common tasks is sleep scoring. 
Various methods have been proposed to address sleep scoring. 
Early studies used Convolutional Neural Networks (CNN)~\cite{supratak2017deepsleepnet, perslev2021u}, Recurrent Neural Networks (RNN)~\citep{phan2019seqsleepnet} for sleep scoring. 
Transformer-based architectures~\cite{phan2022sleeptransformer,lee2024sleepyco,park2025distillsleep} have achieved good predictive performance.
While these works have achieved good sleep scoring results, most of them solely rely on EEG and cannot be used for other downstream tasks than sleep staging. 
Other tasks, such as sleep apnea detection~\cite{levy2023deep, retamales2024towards} and age estimation~\cite{sun2019brain, brink2022age}, have also been proposed, but these works can only be used for certain target tasks and lack generalizability.
Sleep foundation models are proposed to make use of the full potentials included in the multitude of signals from PSG to be used for diverse downstream task. 
\edit{
SleepFM~\cite{thapa2024sleepfm} pioneered this movement to utilize the full range of signals to simultaneously support sleep staging, apnea detection and demographic estimation. 
Its subsequent expansion~\cite{thapa2026multimodal} further demonstrated the utility of large-scale pretraining for clinical disease prediction.
SleepGPT~\cite{huang2026unified} leverages a generative time-frequency fusion approach for staging, pathology classification. 
Sleep2Vec~\cite{yuan2026sleepvec} introduces a subject metadata-aware objective to prevent cohort-specific shortcut learning. 
Meanwhile, OSF~\cite{shuai2026osf} establishes sleep scaling laws and proposes a dual-stage masking strategy to guarantee robustness against missing channels during inference.
Despite these advances, existing sleep foundation models remain fundamentally anchored to localized features (e.g., 30-second epochs) and fail to bridge the structural hierarchy spanning from second-scale fine-grained details to full-night macro-structures.
In contrast, \proposal explicitly unifies both sleep micro- and macro-structures within a single hierarchical framework, achieving superior performance on diverse downstream tasks from localized apnea segmentation to long-term disease prediction.
}

\subsection{Time-series Foundation Model}
Time series data have been studied for decades. 
While statistical models have been mainly used for its analysis, deep learning models recently have shown superior performance.
For more general purpose use across a myriad of different types of data, time-series foundation models have been proposed.
TimesFM~\cite{das2024decoder} proposes decoder-only foundation model pretrained on large corpus of time-series data. 
Moirai~\cite{woo2024unified} proposed cross-frequency and any-variate time series foundation model. 
But these models are limited to time-series forecasting and cannot be directly utilized to sleep downstream tasks, which are mainly classification tasks.
\edit{
To bridge this gap, several representation-learning foundation models have been proposed for classification and general downstream tasks.
MOMENT~\cite{goswami2024moment} and UniTS~\cite{gao2024units} utilize masked autoencoder (MAE) pre-training and task-tokenization to capture multi-dataset patterns across diverse domains. 
More domain-specific models have also been developed: LaBraM~\cite{jiang2024large} trains on massive neural corpora to optimize EEG-specific representations, while HiMAE~\cite{lee2026himae} implements a hierarchical MAE framework tailored for localized wearable sensor streams.
}
While these models could also be used for PSG analysis, its performance is suboptimal as presented in Section~\ref{sec:Experiments}. \proposal has a deep understanding of the characteristics of sleep data, and shows superior performance on various tasks.

\section{\proposal}
\proposal consists of Micro- and Macro- encoders. 
Both encoders are pretrained on large-scale PSG data with self-supervised learning. 
As shown from previous studies~\cite{park2025distillsleep}, PSG scoring inconsistency exists within a dataset and between different datasets due to inter-scorer variability~\cite{danker2004interrater} or guideline updates~\cite{berry2012aasm,berry2017aasm}. 
Independence from scored labels helps \proposal to obtain generalizable embeddings and achieve scalability across diverse datasets.

\subsection{Micro-Encoder}
\textbf{Micro-Encoder architecture}
The Micro-Encoder is designed to map multi-modal raw biosignals into a latent embedding space, enabling the model to capture fine-grained sleep micro-structure. The overall architecture is described in Figure~\ref{fig:micro_encoder}.
Raw signals are first partitioned into a sequence of patches through modality-specific convolutional embedding layers.
These layers accommodate modality-dependent signal characteristics while preserving local temporal dependencies.

The resulting patch sequences are processed by a bifurcated architecture consisting of \textbf{modality-private encoders} and a single \textbf{modality-shared encoder}.
The private encoders specialize in modeling patterns intrinsic to each signal type (e.g., rhythmic oscillations in EEG or QRS complexes in ECG),
whereas the shared encoder focuses on extracting modality-invariant physiological features and cross-modal correlations.

Both private and shared encoders are built upon a Transformer backbone~\cite{vaswani2017attention} enhanced with modern architectural refinements to improve training stability and representational capacity.
Specifically, we adopt rotary positional embeddings~\cite{su2024roformer}, a pre-normalization scheme~\cite{xiong2020layer} using RMSNorm~\cite{zhang2019root}, and SwiGLU activation functions~\cite{shazeer2020glu} within the feed-forward layers.
To further enhance scalability and specialization, the shared encoder employs a Mixture-of-Experts (MoE) architecture~\cite{lepikhin2020gshard}, allowing different experts to selectively model distinct physiological patterns and inter-modal interactions.

The private encoders follow a hierarchical design to enable efficient temporal abstraction.
At lower layers, individual patches are processed independently within each private encoder.
Subsequently, every $M$ consecutive patches are merged into a single higher-level representation, which is then forwarded both to the next private encoder layer and to the shared encoder.
The merged representations from all private encoders are concatenated and jointly processed by the shared encoder, facilitating the modeling of complex inter-modal relationships.
Although concatenation increases the effective input dimensionality with the number of modalities, the hierarchical merging substantially reduces the overall sequence length, maintaining computational efficiency.

To integrate information from the dual streams, the output of the shared encoder is re-partitioned by modality and fused with the corresponding private encoder representations via cross-attention.
The resulting fused embeddings are subsequently up-sampled and passed to a decoder that reconstructs the masked segments of the original biosignals.

The hierarchical merging mechanism requires careful coordination with the masking strategy to preserve temporal alignment.
To ensure that each merged representation corresponds to a consistent temporal span, we employ a structured masking scheme in which masks are applied at fixed intervals.
This design ensures that the reconstruction objective remains well-defined as the model progressively aggregates information across increasing temporal scales.

\textbf{Hybrid Self-supervised Learning} 
The Micro-Encoder is optimized using a synergistic combination of two self-supervised paradigms: Masked Autoencoder (MAE)~\cite{he2022masked} and Contrastive Learning~\cite{oord2018representation, chen2020simple}.
Following the MAE framework, a fixed percentage of patches across all modalities is masked. 
To account for varying dominant frequencies of biosignals included in PSG, we use two different masking patterns. 
For high-frequency biosignals (EEG, EOG, EMG and ECG), we randomly mask a single patch (equivalent to 500ms), and for lower-frequency biosignals (respiratory and oxygen saturation), we randomly mask four consecutive patches (equivalent to 2 seconds). 
The encoders process only the visible patches. 
The latent features are then passed to a decoder to reconstruct the masked portion. 
By minimizing the reconstruction error of the raw signal on masked area, the model learns the fundamental \textit{micro-structure} and local textures of the biosignals.

The reconstruction loss for modality $i$ ($\mathcal{L}_{Mi, (i)}^{recon}$) is formally defined as follows:
\begin{equation}
    \mathcal{L}_{Mi,(i)}^{recon} = \text{MSE}(m_i \cdot x_i, m_i \cdot \hat{x}_i)
\end{equation}
where $x_i$ and $\hat{x}_i$ are the vectors representing original values and the predicted values of modality $i$, and $m_i$ is a mask vector where masked patch area are marked as 1 and 0 otherwise. The loss is averaged over all modalities.

To further refine the latent space, we apply a CL loss to the shared encoder's embeddings. 
This ensures that the shared representations are both temporally consistent and modality-agnostic. 
We define positive pairs as the average of shared embeddings from different modalities sampled within the same timeslot, following the previous works~\cite{thapa2024sleepfm,thapa2026multimodal}.
This encourages the encoder to project different signals representing the same physiological state into a proximal region of the latent space.
Two different kinds of negative pairs are used: temporal negatives and representation negatives. Temporal negative pairs consist of shared embeddings of the same modality from different timeslots. 
Representation negatives pairs consist of a shared embedding and a private embedding from the same signal. 
This specific negative pairing is crucial for feature disentanglement, ensuring the shared encoder does not inadvertently capture modality-specific noise.

The CL loss ($\mathcal{L}_{Mi}^{CL}$) is formally defined as follows:
\begin{equation}
    \mathcal{L}_{Mi, (i,j),t}^{CL} = -  \log\frac{\exp(\langle z_i^t,\bar{z}_{\neq i}^t \rangle /\tau)}{\sum_{s=1}^N \exp( \langle z_i^t, \bar{z}_{\neq i}^s \rangle/\tau) + exp(\langle z_i^t, \tilde{z}^t_i\rangle)}
\end{equation}
where $z_i^t$ is the shared embeddings of modality $i$ at timeslot $t$ and $\bar{z}_{\neq i}^t$ is the average of shared embeddings from all modalities excluding $i$ at $t$. $\tilde{z}_i^t$ is the private embedding of modality $i$ at $t$, $N$ is the total number of timeslots in a batch and $\tau$ is a temperature scaling parameter. KoLeo regularization~\cite{sablayrolles2018spreading} ($\mathcal{L}_{Mi}^{KoLeo}$) is used together for better representation learning.

Final loss is calculated as the sum of reconstruction, contrastive and KoLeo loss as follows:
\begin{equation}
    \mathcal{L}_{Mi} = \mathcal{L}_{Mi}^{recon} + \lambda_{CL} \mathcal{L}_{Mi}^{CL} + \lambda_{KoLeo} \mathcal{L}_{Mi}^{KoLeo}
\end{equation}

\begin{figure}[t]
    \centering
    \includegraphics[width=\columnwidth]{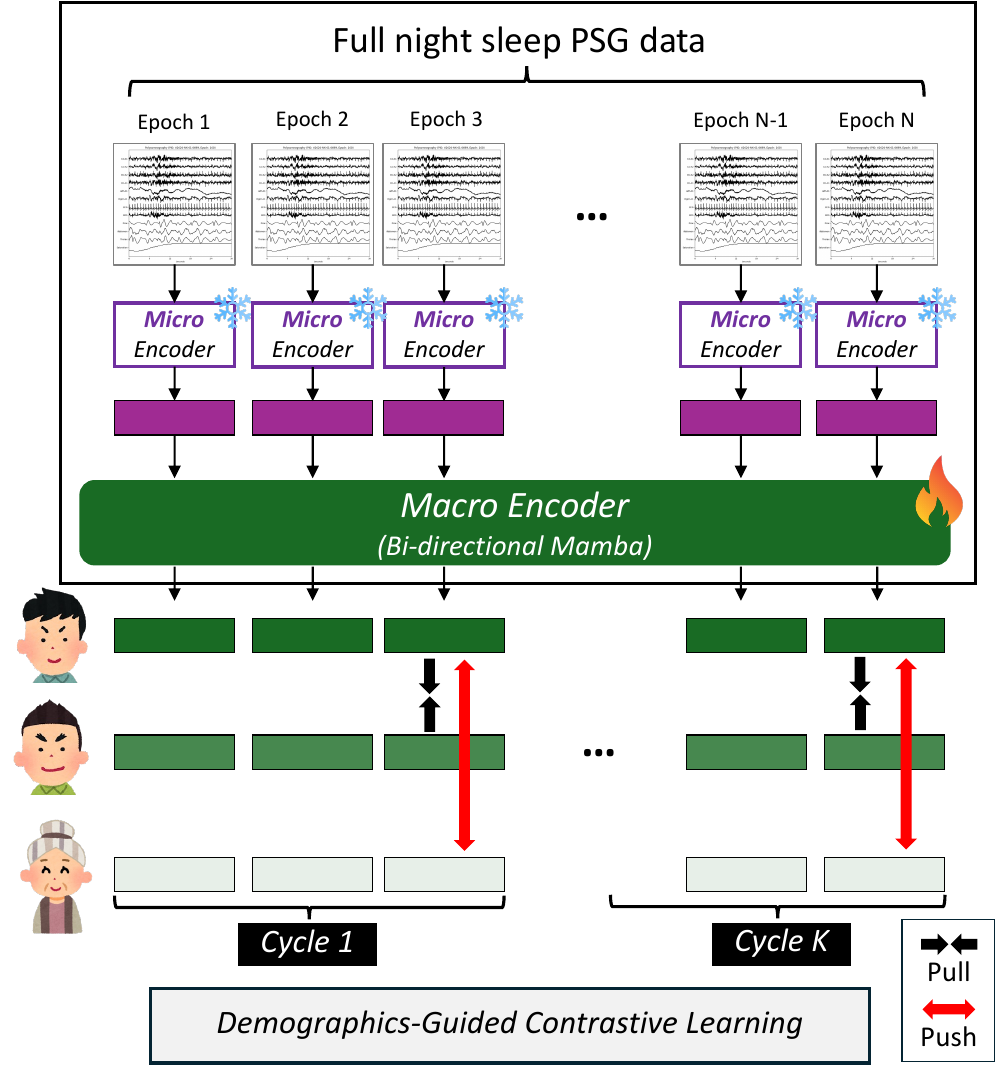}
    \caption{\textbf{Macro-Encoder design and pretraining method.} We utilize bi-directional Mamba layers for efficient long-sequence modeling. Demographic-Guided Contrastive Learning aligns the sleep macro-structure between subjects with objective metadata.}
    \label{fig:macro_encoder}
\end{figure}

\subsection{Macro-Encoder}
The Macro-Encoder is designed to contextualize the local epoch-level embeddings within the broader \textit{macro-structure} of a full night's sleep. 
While the Micro-Encoder captures short-term physiological states, the Macro-Encoder models the temporal evolution of sleep patterns across the entire recording.

\textbf{Macro-Encoder architecture} 
As shown in Figure~\ref{fig:macro_encoder}, the Macro-Encoder receives the complete sequence of epoch-by-epoch embeddings generated by the pretrained Micro-Encoder. 
To process these long-range dependencies, we employ Mamba layers \cite{gu2024mamba}. 
We specifically choose the Mamba architecture (a Selective State Space Model) over traditional Transformers due to its superior scaling properties and memory efficiency. 
Given that a standard full-night recording typically exceeds 800 epochs and large batch size expedites for effective contrastive learning used for training, Mamba’s linear scaling allows it to capture global context without the quadratic memory overhead associated with self-attention.
Furthermore, we implement these layers in a bi-directional manner to address the inherent variability in recording durations across subjects, making fixed-length temporal alignment difficult especially at later stages. 
Bi-directionality allows the model to learn the macro-structure both from the sleep onset and from the sleep termination.

\begin{figure}[t]
    \centering
    \begin{subfigure}[b]{\columnwidth}
        \centering
        \includegraphics[width=0.9\textwidth, bb=0 0 679 423]{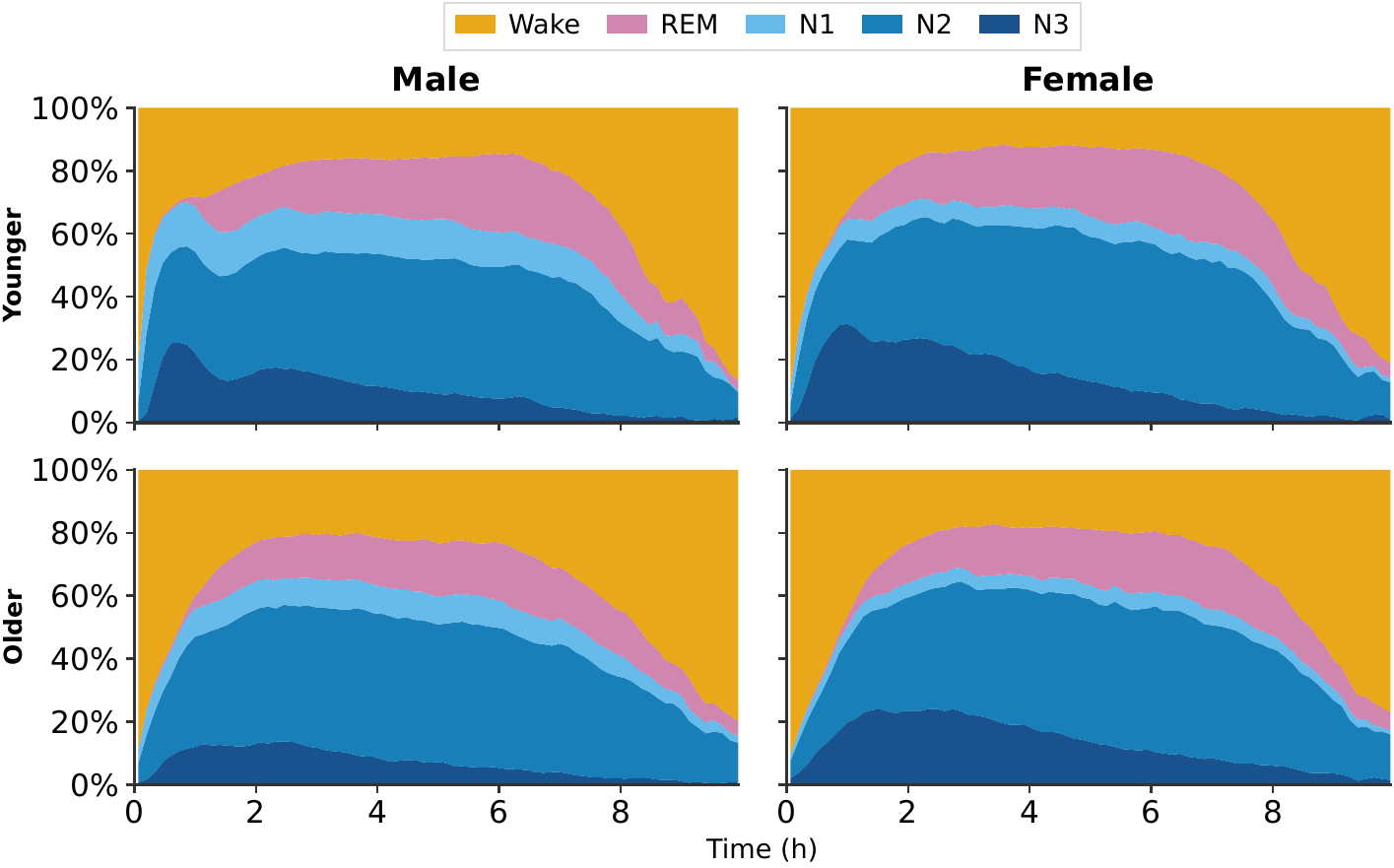}
        \vspace{-1ex}
        \caption{Non-obese (BMI $<$ 30 kg/m$^2$)}
        \label{fig:sleep_nonobese}
    \end{subfigure}
    
    \vspace{0.3cm}
    
    \begin{subfigure}[b]{\columnwidth}
        \centering
        \includegraphics[width=0.9\textwidth, bb=0 0 679 423]{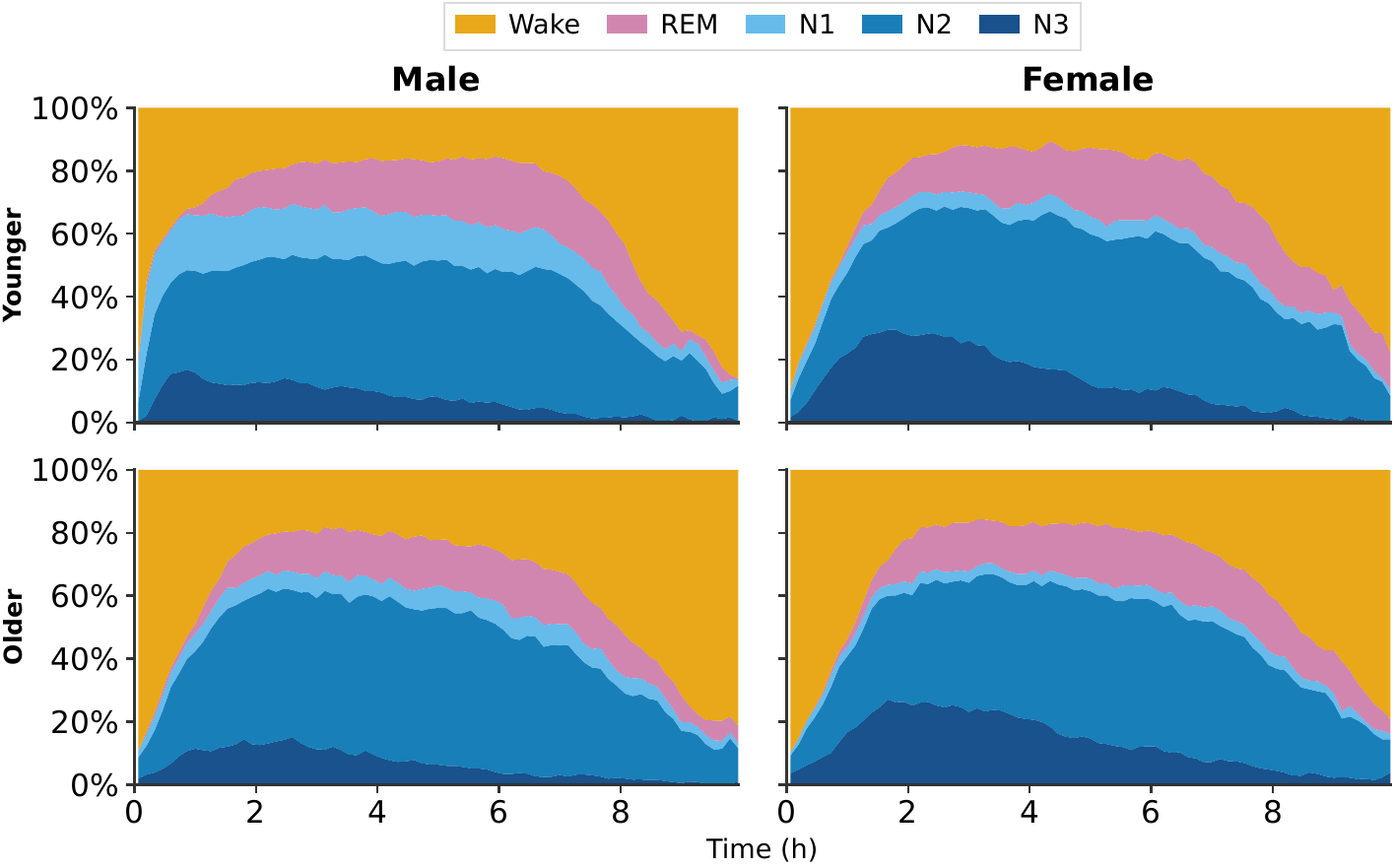}
        \vspace{-1ex}
        \caption{Obese (BMI $\geq$ 30 kg/m$^2$)}
        \label{fig:sleep_obese}
    \end{subfigure}
    \caption{\textbf{Sleep macro-structure variations across demographic groups}. Sleep stage distributions over full-night recordings by sex, age (Younger: $<$ 60 yrs; Older: $\geq$ 60 yrs) and BMI. N3 proportion in the early sleep period or REM sleep proportion in later stage varies significantly across groups. These demographic dependent patterns motivate our Demographic-Guided Contrastive Learning objective.}
    \label{fig:sleep_macrostructure}
    \vspace{-2ex}
\end{figure}

\textbf{Demographic-guided Contrastive Learning (DGCL)} To optimize the Macro-Encoder, we introduce a variant of contrastive learning that leverages demographic metadata, specifically age, sex and BMI as supervisory signals. 
These factors are primary determinants of sleep architecture, influencing parameters such as Slow Wave Sleep (SWS) or REM sleep duration and sleep fragmentation \cite{mander2017sleep}.
Figure~\ref{fig:sleep_macrostructure} shows the sleep stage distribution trend over full night, separated by age (Younger (18-60 yrs) and Older ($>$ 60 yrs)), BMI (Non-obese and Obese) and sex, derived from a combined subset of our pretraining data (SHHS1/2, KISS and KVSS).
The distributions reveal clear distinction between different groups. 
For example, the N3 proportion in the older, obese male group  (bottom left) is significantly lower than that of younger, non-obese female group (top right) during the early stages of sleep. In addition, REM proportion in the later stages of the night vary considerably across different groups. 
These demographic attributes provide objective, noise-free ground truth, circumventing inter-scorer variability~\cite{danker2004interrater,danker2009interrater} and evolving clinical guidelines~\cite{berry2012aasm,berry2017aasm} associated with manual PSG labeling.
Furthermore, the near-universal availability of demographic data ensures the scalability of this training strategy.
\edit{More visualizations are provided in Appendix~\ref{sec:more_macro_analysis}.}

To effectively capture sleep cycles, we partition the full-night sequence into 90-minute intervals (180 epochs), corresponding to the average duration of a human ultradian sleep cycle~\cite{dement1957cyclic}. 
We then apply a soft-target contrastive objective that calculates the similarity between patient intervals based on their demographic profiles. 
Instead of traditional binary positive/negative pairs, we adopt a weighted similarity approach, inspired by generalized InfoNCE loss~\cite{yang2023simper}. 
For a given interval $c$ and subject pair $(i,j)$, the loss for forward Mamba pass is formulated as follows:
\begin{equation}
    \begin{aligned}
        \overrightarrow{\mathcal{L}}_{Ma, (i,j),c} &= - w_{i,j} \log \frac{\exp(\langle \overrightarrow{Z}_i^c,\overrightarrow{Z}_j^c \rangle /\rho)}
        {\sum_{k=1}^{K} \exp( \langle \overrightarrow{Z}_i^c, \overrightarrow{Z}_k^c \rangle/\rho)} \\
        w_{i,j}& = \frac{\exp(-d_{i,j}/\upsilon)}{\sum_{k=1}^{K} \exp(-d_{i,k}/\upsilon)}         
    \end{aligned}
\end{equation}
where $\overrightarrow{Z}_{i}^c$ and $\overrightarrow{Z}_{j}^c$ are the refined latent features of the subjects $i$ and $j$ at the end of $c$-th interval in the forward pass, $\rho$ and $\upsilon$ are temperature scaling parameters and $K$ is the total number of subjects in a batch. 
The \textit{demographic distance}$ d_{i,j}$ serves as the supervisory signal, calculated as a function of the mean absolute difference of age and BMI, with a penalty for sex mismatch: 
\begin{equation}
    \begin{aligned}                
        d_{i,j} &= (|\text{age}_i - \text{age}_j| + |\text{BMI}_i - \text{BMI}_j|)/2 +\lambda_{sex} 
    \end{aligned}
\end{equation}
Here, age and BMI are z-score normalized and $\lambda_{sex}$ is a constant penalty applied only when subjects $i$ and $j$ are of different sex. 
The backward loss ($\overleftarrow{\mathcal{L}}_{Ma, (i,j),c}$) is defined symmetrically for the backward Mamba pass. The total Macro loss ($\mathcal{L}_{Ma}$) is aggregated over all intervals and subject pairs:
\begin{equation}
    \mathcal{L}_{Ma} = \sum_c\sum_{i \neq j}  \left( \overrightarrow{\mathcal{L}}_{Ma, (i,j),c} + \overleftarrow{\mathcal{L}}_{Ma, (i,j),c} \right)    
\end{equation}
This strategy effectively regularizes the latent space by pulling subjects with similar demographic profiles closer together while pushing disparate subjects further apart. 
To ensure stable convergence, we employ a stratified batch sampling strategy, maintaining balanced demographic distributions and similar sequence lengths within each batch.

\subsection{Pretraining datasets and preprocessing}
To pretrain \proposal, we utilize the combination of open-sourced sleep datasets: the Sleep Heart Health Study (SHHS1/2)~\cite{zhang2018national, quan1997sleep}, Korea Image-based Sleep Study (KISS)~\cite{jeong2023standardized}, Korea Video Sleep Study (KVSS)~\cite{choi2024kvss}, Physionet 2018 (PHY)~\cite{goldberger2000physiobank, ghassemi2018you}, Multi-Ethnic Study of Atherosclerosis (MESA)~\cite{chen2015racial} and Osteoporotic Fractures in Men Study (MrOS)~\cite{blackwell2011associations}. 
The training split of SHHS1 and KISS are used for pretraining, with the partitioning scheme following the established protocols~\cite{phan2021xsleepnet, park2025distillsleep}. In total, we use 20,964 PSG recordings (158,028 hours) for pretraining. 
\edit{More details on datasets are provided in Appendix~\ref{sec:dataset}}.

\edit{
We use biosignals from 6 different modalities: electroencephalogram (EEG), electrooculogram (EOG), electrocardiogram (ECG), electromyogram (EMG), respiratory signals (Nasal airflow, Abdominal/Thoracic) and $\text{SpO}_2$.
}
All raw biosignals are resampled at 100 Hz to standardize the varying sampling rates across the diverse datasets. We applied a 0.3–40 Hz bandpass filter to the EEG and EOG signals, a 0.3–60 Hz bandpass filter to the ECG, and a 10 Hz highpass filter to the EMG. 
Additionally, a 60 Hz notch filter was applied to the EEG, EOG, ECG, and EMG signals to remove power-line interference.
Respiratory signals were processed with a 15 Hz lowpass filter. 
Finally, all signals \edit{except $\text{SpO}_2$} were z-score normalized   prior to model input. 

\subsection{Implementation}
The code is written in PyTorch and PyTorch Lightning wrapper. Both Macro- and Micro-Encoders are trained with AdamW~\cite{loshchilov2018decoupled} optimizer with cosine annealing~\cite{loshchilov2017sgdr} learning rate schedule. The training is done on 2 $\times$ NVIDIA H100 HBM3 (VRAM 80GB) GPUs and the end-to-end pretraining takes approximately 10 hours. Flash attention-2~\cite{dao2023flashattention} is utilized in Transformer blocks for efficiency. More training details are provided in Appendix.~\ref{sec:training_details}.

\section{Experiments and Results} \label{sec:Experiments}

\subsection{Downstream Tasks}
We evaluate \proposal on downstream tasks including sleep stage classification, sleep disordered breathing (SDB) segmentation and disease prediction. 
To compare the performance of \proposal to other contemporary foundation models we choose two time series foundation models (MOMENT~\cite{goswami2024moment} and UniTS~\cite{gao2024units}), a EEG foundation model (LaBraM~\cite{jiang2024large}) and a sleep foundation model (SleepFM-Disease~\cite{thapa2026multimodal}) as baselines. 
Unless the baseline model's input format requires specific settings (e.g. Sampling rate of 256 Hz for SleepFM-Disease), we use the same input with identical preprocessing.
\edit{Except disease prediction, the performance is measured on the test split of SHHS1 and KISS as well as two held-out datasets; Cleveland Family Study (CFS)~\cite{zhang2018national, redline1995familial} and Study of Osteoporotic Fractures (SOF)~\cite{zhang2018national, spira2008sleep}.  Disease prediction performance is only evaluated on SHHS1 where disease history is provided with PSG data.}

\subsubsection{Sleep stage classification}
Sleep staging refers to classifying each 30-second epoch into one of five sleep stages: Wake, Rapid Eye Movement (REM), Non-REM stages N1, N2 and N3. We conduct linear probing~\cite{alain2017understanding} on target datasets to objectively test the quality of the learned representations. Weighted cross entropy loss is used to account for imbalanced sleep stage distribution. In addition, following conventional sleep scoring protocols, we truncated continuous Wake epochs to 30 minutes before the first non-Wake epoch and 30 minutes after the last non-Wake epoch. 

The sleep staging results are provided in Table~\ref{tab:sleep_staging}. \edit{On all evaluation datasets, \proposal outperforms all baselines, demonstrating its superior capability in capturing sleep-specific physiological features. 
Especially, \proposal achieved 80.7\% of accuracy and 71.2\% of Macro-F1 on CFS, 79.1\% of accuracy and 65.2\% of Macro-F1 on SOF, showing its generalizability to unseen datasets. Moreover, while KISS dataset was collected from different PSG systems (Nox and Embla) (see Table~\ref{tab:channels} in Appendix), \proposal shows robust sleep staging performance on this dataset.
}


\begin{table}[t]
    \centering
    \caption{\textbf{Sleep stage classification results.} The results demonstrate the effectiveness of \proposal compared to time series and sleep foundation models. The evaluation is done on the test split of SHHS1, KISS datasets, \edit{and held-out datasets(CFS, SOF)}}
    
    \label{tab:sleep_staging}
    \resizebox{0.95\linewidth}{!}{%
    \begin{tabular}{cccccc}
        \toprule
        \textbf{Dataset} & \textbf{Category} & \textbf{Models} & \textbf{Accuracy} & \textbf{Macro-F1} & \textbf{Kappa} \\ 
        \midrule
        \multirow{5}{*}{SHHS1} & Time series & MOMENT-Base & 79.4 & 65.6 & \textbf{70.0} \\
         & Foundation model & UniTS & 64.2 & 59.2 & 53.3 \\ 
         \cmidrule(lr){2-6}
         & \edit{EEG FM} & \edit{LaBraM} & \edit{70.2} & \edit{54.1} & \edit{55.7} \\
         \cmidrule(lr){2-6}
         & \multirow{2}{*}{\begin{tabular}[c]{@{}c@{}}Sleep\\ Foundation model\end{tabular}} & SleepFM-Disease & 69.7 & 56.3 & 55.9 \\
         &  & \textbf{\proposal} (Ours) & \textbf{81.9} & \textbf{74.1} & \textbf{70.0} \\ 
        \midrule
        \multirow{5}{*}{KISS} & Time series & MOMENT-Base & 69.8 & 66.3 & 59.7 \\
         & Foundation model & UniTS & 60.3 & 58.8 & 49.8 \\ 
         \cmidrule(lr){2-6}
        & \edit{EEG FM} & \edit{LaBraM} & \edit{58.7} & \edit{50.1} & \edit{43.2} \\
         \cmidrule(lr){2-6}
         
         & \multirow{2}{*}{\begin{tabular}[c]{@{}c@{}}Sleep\\ Foundation model\end{tabular}} & SleepFM-Disease & 57.6 & 58.4 & 46.1 \\
         &  & \textbf{\proposal} (Ours) & \edit{\textbf{71.0}} & \edit{\textbf{70.0}} & \edit{\textbf{62.0}} \\ 
        \midrule
         \multirow{4}{*}{\edit{CFS}} & \edit{Time series} & \edit{MOMENT-Base} & \edit{71.4} & \edit{52.4} & \edit{56.7} \\
         & \edit{Foundation model} & \edit{UniTS} & \edit{64.1} & \edit{58.1} & \edit{53.3} \\ \cmidrule(lr){2-6}
         & \multirow{2}{*}{\edit{\begin{tabular}[c]{@{}c@{}}Sleep\\ Foundation model\end{tabular}}} & \edit{SleepFM-Disease} & \edit{71.1} & \edit{57.4} & \edit{57.5} \\
         &  & \edit{\textbf{\proposal} (Ours)} & \edit{\textbf{80.7}} & \edit{\textbf{71.2}} & \edit{\textbf{72.9}} \\

        \midrule
         \multirow{4}{*}{\edit{SOF}} & \edit{Time series} & \edit{MOMENT-Base} & \edit{79.0} & \edit{62.6} & \edit{70.0} \\
         & \edit{Foundation model} & \edit{UniTS} & \edit{76.1} & \edit{60.5} & \edit{65.9} \\ \cmidrule(lr){2-6}
         & \multirow{2}{*}{\edit{\begin{tabular}[c]{@{}c@{}}Sleep\\ Foundation model\end{tabular}}} & \edit{SleepFM-Disease} & \edit{60.1} & \edit{47.0} & \edit{42.3} \\
         &  & \edit{\textbf{\proposal} (Ours)} & \edit{\textbf{79.1}} & \edit{\textbf{65.2}} & \edit{\textbf{70.4}} \\
         
        \bottomrule
    \end{tabular}
    }
    \vspace{-1ex}
\end{table}

\begin{table}[t]
    \centering
    \caption{\textbf{SDB segmentation results.} The results demonstrate \proposal's superior capability to capture fine-grained details, compared to time series and sleep foundation models. The evaluation is done on the test split of SHHS1 and KISS datasets. F1 is emphasized as a more robust metric under severe class imbalance in SDB labels.}
    
    \label{tab:sleep_apnea_seg}
    \resizebox{0.98\linewidth}{!}{%
    \begin{tabular}{ccccc}
        \toprule
        \textbf{Dataset} & \textbf{Category} & \textbf{Models} & \textbf{Accuracy} & \textbf{F1} \\ 
        \midrule
        \multirow{5}{*}{SHHS1} & \multirow{1}{*}{Time series} & MOMENT-Base & 73.4 & 33.4  \\
         &  \multirow{1}{*}{Foundation model} & UniTS & 88.2 & \textbf{48.8}  \\ \cmidrule(lr){2-5}
         & \edit{EEG FM} & \edit{LaBraM} & \edit{50.2} & \edit{20.4} \\
         \cmidrule(lr){2-5}
         & \multirow{1}{*}{Sleep} & SleepFM-Disease & 77.5 & 39.4  \\
         & \multirow{1}{*}{Foundation model} & \textbf{\proposal} (Ours) & \textbf{88.4} & 48.5  \\ 
        \midrule
        \multirow{5}{*}{KISS} & \multirow{1}{*}{Time series} & MOMENT-Base & 76.0 & 58.5  \\
         & \multirow{1}{*}{Foundation model} & UniTS & 79.8 & 63.6  \\ 
         \cmidrule(lr){2-5}
         & \edit{EEG FM} & \edit{LaBraM} & \edit{56.3} & \edit{38.5} \\
         \cmidrule(lr){2-5}
         & \multirow{1}{*}{Sleep} & SleepFM-Disease & 79.6 & 65.1 \\
         & \multirow{1}{*}{Foundation model} & \textbf{\proposal} (Ours) & \edit{\textbf{83.7}} & \edit{\textbf{66.2}}  \\ 

         \midrule
        \multirow{4}{*}{\edit{CFS}} & \multirow{1}{*}{Time series} & MOMENT-Base & 74.2 & 38.5  \\
         & \multirow{1}{*}{Foundation model} & UniTS & 85.5 & 40.0  \\ \cmidrule(lr){2-5}
         & \multirow{1}{*}{Sleep} & SleepFM-Disease & 82.1 & 46.6 \\
         & \multirow{1}{*}{Foundation model} & \textbf{\proposal} (Ours) & \edit{\textbf{88.7}} & \edit{\textbf{52.3}}  \\ 

         \midrule
        \multirow{4}{*}{\edit{SOF}} & \multirow{1}{*}{Time series} & MOMENT-Base & 70.3 & 18.3  \\
         & \multirow{1}{*}{Foundation model} & UniTS & \textbf{90.6} & \textbf{35.7}  \\ \cmidrule(lr){2-5}
         & \multirow{1}{*}{Sleep} & SleepFM-Disease & 79.1 & 30.1 \\
         & \multirow{1}{*}{Foundation model} & \textbf{\proposal} (Ours) & \edit{89.4} & \edit{32.7}  \\ 
        \bottomrule
    \end{tabular}
    }
    \begin{flushleft}
        \scriptsize *The ICML version of this table reported Macro-F1 for \proposal, but F1 of sleep disoredred breathing class for baselines. This version reports F1 of sleep disordered breathing class for all models for consistency.
    \end{flushleft}
    \vspace{-2ex}
\end{table}

\subsubsection{SDB segmentation}
SDB segmentation requires the high-resolution classification of each one-second interval into categories of normal breathing or disordered breathing (specifically hypopnea or apnea). 
Consistent with our evaluation protocol for sleep stage classification, we evaluate the learned representations via a linear probing task optimized with weighted cross entropy loss. 

The results are summarized in Table~\ref{tab:sleep_apnea_seg}. We present the performance of \proposal using accuracy and F1-score for disordered breathing.
\proposal achieves the highest accuracy on SHHS1, KISS, and CFS, and the highest F1-score on KISS and CFS.
On SOF, UniTS achieves the best results on both metrics; \proposal nevertheless performs competitively, with accuracy close to that of UniTS.
The results also show consistent gains over SleepFM-Disease across all four datasets. We hypothesize that these gains are attributed to \proposal's hybrid pretraining strategy. SleepFM solely relies on contrastive learning (CL), which emphasizes global signal alignment, whereas \proposal combines CL with masked autoencoder (MAE) reconstruction. The reconstruction objective may encourage the model to capture local respiratory signal structure, which is useful for detecting subtle breathing disruptions at one-second resolution.

SOF remains a more challenging dataset for \proposal. Its older participant population and distinct PSG acquisition protocols may contribute to the comparatively weaker results, although further analysis is needed to establish the cause. Improving generalization to such datasets is an important direction for future work.


\subsubsection{Disease Prediction}
We utilize the SHHS dataset's disease histories to perform PSG-based disease prediction. We also employ linear probing for this task using the Cox Proportional Hazard (Cox PH) loss~\citep{cox1972regression, katzman2018deepsurv}, defined by the following negative log-partial likelihood:
\begin{equation}
    \mathcal{L}_{PH} = \frac{1}{N_e} \sum_{i=1}^{N} -\delta_i \left( h_i - \log \sum_{k \in R_i} \exp(h_k) \right)
\end{equation}
where $h_i$ represents the predicted risk score for subject $i$, and $\delta_i \in \{0,1\}$ is a binary indicator denoting whether subject $i$ experienced the event (disease occurrence). $N_e=\sum_i \delta_i$ is the total number of the events within the population of $N$ subjects. 
The risk set $R_i$ consists of all subjects whose time-to-event (or censoring) is greater than or equal to that of subject $i$. 
Following the evaluation protocol of SleepFM-Disease~\cite{thapa2026multimodal}, we assess the predictive performance of \proposal across six diseases. 
Performance is quantified via Harrell’s concordance index (C-Index).
The results are summarized in Table~\ref{tab:disease_results}. \proposal shows comparable performance to SleepFM-Disease, highlighting its potential for reliable clinical application in sleep-based healthcare.

\begin{table}[t]
    \centering
    \caption{\textbf{Disease prediction.} C-Index is reported on 6 selected diseases from SHHS1 dataset. \proposal shows comparable performance to the state-of-the-art PSG-based disease prediction model.}
    \label{tab:disease_results}
    \resizebox{\linewidth}{!}{%
    \begin{tabular}{@{}ccccccc@{}}
    \toprule
        \multirow{2}{*}{\textbf{Models}} & \multicolumn{6}{c}{\textbf{Disease Outcomes}} \\ \cmidrule(l){2-7} 
         & \textbf{Angina} & \textbf{CVD death} & \textbf{CHF} & \textbf{CHD death} & \textbf{MI} & \textbf{Stroke} \\ \midrule
        \multicolumn{7}{c}{\textit{C-Index}} \\ \midrule
        SleepFM-Disease & 0.632 & \textbf{0.791} & 0.764 & \textbf{0.781} & 0.636 & \textbf{0.729} \\
        \proposal (Ours) & \textbf{0.778} & 0.788 & \textbf{0.793} & 0.776 & \textbf{0.662} & 0.718 \\ \midrule
    \end{tabular}%
    }
    \begin{flushleft}
        \scriptsize *CVD death: CardioVascular Disease death, CHF: Congestive Heart Failure, CHD death: Coronary Heart Disease death, MI: Myocardial Infarctions
    \end{flushleft}
    \vspace{-2ex}
\end{table}

\subsection{Few-shot Evaluation}

\begin{figure*}[t]
    \centering
    \begin{subfigure}[b]{0.3\textwidth}
        \centering
        \includegraphics[width=\textwidth, bb=0 0 215 211]{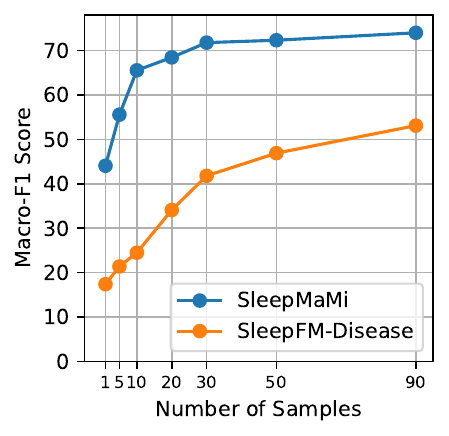}
        \vspace{-2ex}
        \caption{Sleep staging (ISRUC)}
        \label{fig:fewshot_f1}
    \end{subfigure}
    \begin{subfigure}[b]{0.3\textwidth}
        \centering
        \includegraphics[width=\textwidth, bb=0 0 215 211]{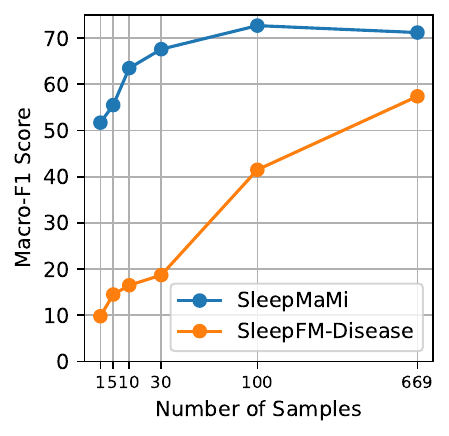}
        \vspace{-2ex}
        \caption{\edit{Sleep staging (CFS)}}
        \label{fig:fewshot_f1}
    \end{subfigure}
    \begin{subfigure}[b]{0.3\textwidth}
        \centering
        \includegraphics[width=\textwidth, bb=0 0 215 211]{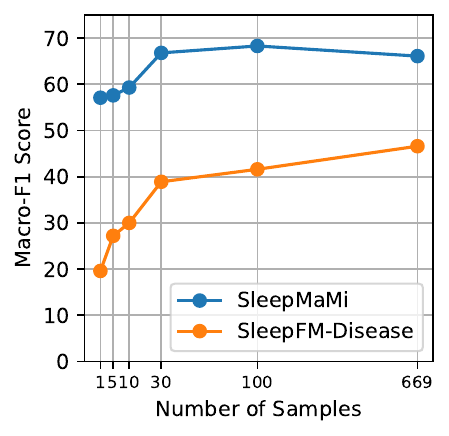}
        \vspace{-2ex}
        \caption{\edit{SDB segmentation (CFS)}}
        \label{fig:fewshot_f1}
    \end{subfigure}
    \caption{\textbf{Few shot evaluation.} The performance is measured with varying number of fine-tuning samples in Macro-F1 score. \proposal shows label-efficient adaptation to unseen dataset thanks to well-generalized embeddings derived from large-scale pretraining data.}
    \label{fig:fewshot}
\end{figure*}

To evaluate the fast adaptation ability of \proposal to unseen data, we conduct a few-shot evaluation \edit{using the ISRUC and CFS datasets~\cite{khalighi2016isruc, redline1995familial}. For ISRUC, Subgroup 1 is utilized for fine-tuning and validation, while Subgroups 2 and 3 are combined to form the test set. 
We restrict the fine-tuning data from Subgroup 1 to specific increments ($n = 1, 5, 10, 20, 30, 50, 90\text{ (All)}$) to evaluate the sleep staging performance on the test split.
For CFS, we use a varying number of samples ($n = 1, 5, 10, 30, 100, 669  \text{ (All)}$) from the training split for finetuning, evaluating  sleep staging and SDB segmentation performance on the test split. 
As illustrated in Figure \ref{fig:fewshot}, \proposal demonstrates impressive label efficiency. 
With only a single training sample, the model achieves F1 scores of 44.1\%, 51.7\%, and 57.1\% for ISRUC sleep staging, CFS sleep staging, and CFS SDB segmentation, respectively.
When the sample size increases to 30, the Macro-F1 scores rise to 71.8\%, 67.6\%, and 66.8\%, which are only 2.2\%, 3.6\%, and 1.5\% lower than the highest Macro-F1 scores. 
These results validate the utility of \proposal as a robust foundation model capable of generalizing in data-constrained scenarios.
}

\begin{table}[t]
    \centering
    \caption{\textbf{Performance comparison between Micro-only and \proposal.} The integration of the Macro-Encoder consistently improves performance across all tasks, validating the effectiveness of DGCL in incorporating sleep macro-structure.}
    \label{tab:macro_micro_comparison}
    \resizebox{\linewidth}{!}{%
    \vspace{-1ex}
    \begin{tabular}{@{}lccccc@{}}
        \toprule
        \multirow{2}{*}{\textbf{Models}} & \multicolumn{2}{c}{\textbf{Classification (Acc. $\uparrow$)}} & \multicolumn{2}{c}{\textbf{Regression (MAE $\downarrow$)}} & \textbf{\edit{Disease}} \\ 
        \cmidrule(lr){2-3} \cmidrule(l){4-5} & Sleep stage & Sex & Age & AHI & \textbf{\edit{Prediction}} \\ \midrule
        Micro-Encoder only & 79.8 & 80.1 & 9.45 & 9.69 & \edit{0.729}\\
        \textbf{\proposal} & \textbf{81.9} & \textbf{88.2} & \textbf{6.73} & \textbf{8.40} & \textbf{\edit{0.753}} \\ \bottomrule
    \end{tabular}    
    }
    \vspace{-3ex}
\end{table}

\subsection{Contribution of the Macro-Encoder}

Unlike existing models that focus exclusively on sleep micro-structure, \proposal incorporates a Macro-Encoder to integrate high-level sleep \textit{macro-structure} with micro-structure.
To evaluate the added value of the Macro-Encoder, we conduct a comparative analysis using two classification, two regression tasks \edit{and disease prediction task}, evaluating embeddings from the Micro-Encoder alone against those from the full \proposal architecture. 
As summarized in Table~\ref{tab:macro_micro_comparison}, the integration of the Macro-Encoder yielded performance gains across all \edit{five} tasks. 
\edit{
Notably, while demographic tasks naturally benefit from the metadata used during pre-training, the Macro-Encoder drives substantial performance gains in independent clinical downstream tasks, including sleep staging, AHI estimation, and disease prediction. 
}
These results suggest that DGCL fosters a superior representation of the global sleep architecture, benefiting a wide range of downstream sleep-related tasks.

\subsection{Embedding Analysis and Visualization}

\begin{figure}[t]
    \centering
    \begin{subfigure}[b]{0.49\columnwidth}
        \centering        
        \includegraphics[width=0.99\textwidth]{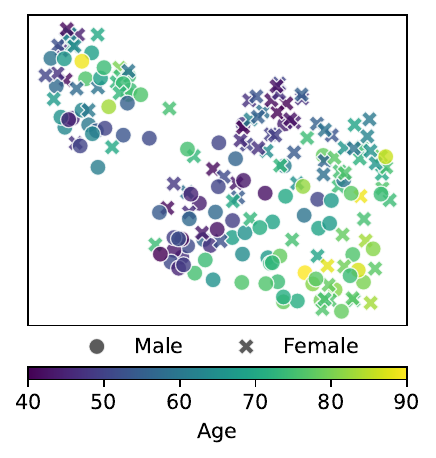}
        \caption{SHHS1}
        \label{fig:umap_shhs}
    \end{subfigure}    
    \begin{subfigure}[b]{0.49\columnwidth}
        \centering
        \includegraphics[width=0.99\textwidth]{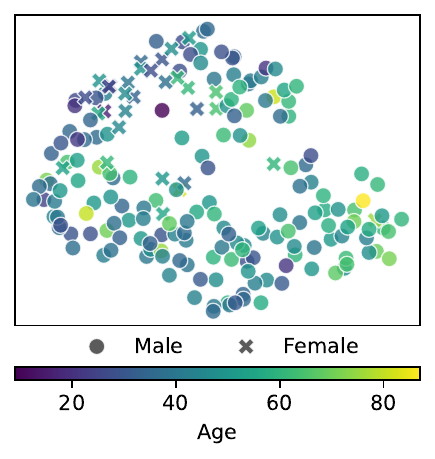}
        \caption{KISS}
        \label{fig:umap_kiss}
    \end{subfigure}
    \caption{\textbf{Macro-embeddings visualization.} Per-subject latent embeddings from Macro-Encoder is visualized using U-MAP. Each point represents a single subject's embeddings where color and symbol represents age and sex, respectively. More visualizations are available in Appendix.~\ref{sec:more_viz}}
    \label{fig:umap}
    \vspace{-3ex}
\end{figure}

To qualitatively assess how effectively the Macro-Encoder captures demographic features within the latent space, we visualize the subject-level embeddings using U-MAP, as shown in Figure~\ref{fig:umap}. 
This analysis is performed on 200 subjects randomly sampled from the test splits of the SHHS1 and KISS datasets. 
The visualization reveals clear clustering based on demographic attributes, with subjects of similar age and sex residing closely within the embedding space. 
This distinct separation confirms that our DGCL framework successfully maps demographic characteristics into the latent space, providing the model with a grounded, subject-aware context for sleep analysis. 
\edit{More embedding visualizations are provided in Appendix~\ref{sec:more_viz}}

\subsection{More experimental results}

\edit{
Additional experimental results are provided in Appendix~\ref{sec:more_experiments}. Specifically, Table~\ref{tab:ablation_study} presents the ablation study of \proposal's core components, while Tables~\ref{tab:demographic_ablation} and~\ref{tab:cycle_len_ablation} detail the ablation results regarding demographic factors and the cycle lengths used in DGCL, respectively. Furthermore, Table~\ref{tab:macro_encoder_ablation} provides the ablation results on the Macro-Encoder architecture, and Table~\ref{tab:macro_encoder_pretrain_ablation} compares DGCL against alternative macro-level pretraining strategies. Finally, Tables~\ref{tab:sleep_staging_lstm} and~\ref{tab:disease_prediction_lstm} report performance under different evaluation protocols when using an LSTM classifier.
}

\section{Conclusion}

In this study, we introduce \proposal, a novel sleep foundation model that integrates both sleep \textit{micro-} and \textit{macro-structure}, pretrained on 20,964 PSG recordings. 
Our architecture utilizes a dual-encoder design: a Micro-Encoder with a private-shared transformer backbone optimized via hybrid of MAE and CL to capture the fine-grained physiological features and a Mamba-based Macro-Encoder to model long-range temporal dependencies across full night. 
DGCL is employed to train the Macro-Encoder, which leverages objective metadata to encode sleep macro-structure into the model. 
Extensive evaluations demonstrate that \proposal outperforms existing foundation models across a wide spectrum of tasks. Furthermore, the model exhibits strong label-efficient adaptation, proving its utility as a robust and practical foundation for clinical sleep analysis even in data-constrained environment. 
\section*{Acknowledgement}

\edit{
This work was supported in part by the Institute of Information \& Communications Technology Planning \& Evaluation (IITP) grant funded by the Korea government (MSIT) (RS-2025-25463302, Global Talent Recruitment for AI-powered Drug Discovery), 
in part by the National Research Foundation (NRF) of Korea grant funded by the Korea government (MSIT)
(No. RS-2023-00222663),
in part by the AI-Bio Research Grant through Seoul National University,
in part by the AI Computing Infrastructure Enhancement (GPU Rental Support) User Support Program funded by the Ministry of Science and ICT (MSIT), in part by the “Advanced GPU Utilization Support Program” funded by the Government of the Republic of Korea (Ministry of Science and ICT).
We also gratefully acknowledge Heonjun Lee's support for obtaining experimental results.
}

\section*{Impact Statement}
\edit{
Our work introduces a universal sleep foundation model that advances automated polysomnography (PSG) analysis by explicitly unifying localized micro-signal morphologies with global, full-night macro-structural architectures. 
By establishing a robust pretraining framework that achieves high performance entirely independent of manual annotations, the model effectively mitigates the long-standing challenges of severe inter-scorer variability and shifting clinical scoring guidelines. 
This provides a highly standardized baseline for automated sleep medicine, paving the way for more reliable, scalable, and equitable diagnostic tools.  
Furthermore, the model demonstrates impressive label efficiency, adapting rapidly to unseen datasets and clinical downstream tasks with minimal fine-tuning data. 
This fast adaptation capability is particularly vital for resource-constrained clinical settings, rare sleep disorder research, and emerging diagnostic paradigms where massive annotated medical data is difficult to acquire.  
Unlike conventional models limited to localized epoch dynamics, \proposal captures long-range temporal dependencies across an entire full-night recording. 
This holistic representation opens critical new avenues for non-invasive, long-term prognostic disease prediction and preventive medicine.  
}

\edit{
While our framework demonstrates significant promise, we recognize that the deployment of artificial intelligence in clinical settings demands the highest standards of reliability, transparency, and fairness. 
Because \proposal utilizes demographic metadata within the Demographic-Guided Contrastive Learning (DGCL) framework to organize the latent space, there is an inherent risk of capturing or perpetuating historical clinical biases. 
Although our empirical subgroup analysis demonstrates that the model maintains highly robust and fair performance across diverse cohorts without direct input dependencies, proactively neutralizing potential demographic and demographic-shortcut biases remains a critical priority for future research. 
As deep learning methodologies increasingly intersect with clinical sleep medicine, ensuring robust ethical safeguards around data consent, model explainability, and algorithmic accountability is essential. 
By prioritizing accessibility, technical responsibility, and broader social alignment, we believe these technologies can safely and effectively transform clinical workflows. 
Ultimately, this work represents a meaningful step toward leveraging foundation models to unlock the multi-modal richness and macro-structural characteristics of physiological sleep data, providing a scalable pathway to deepen our understanding of systemic health.
}

\section*{Conflict of Interest}
\edit{
H.-W.S. is an inventor on patent applications submitted by Seoul National University related to an image-based polysomnography dataset and its application. H.-W.S. is a founder of OUaR LaB, Inc., serves on the Board of Directors and as a chief executive officer for OUaR LaB, Inc., and owns OUaR LaB Stock, which are subject to certain restrictions under university policy. All other authors declare no competing interests. 
}


\bibliography{main}
\bibliographystyle{icml2026}

\newpage
\appendix
\onecolumn


\section{Datasets} \label{sec:dataset}
The brief description of the datasets used for this research is provided below. Table~\ref{tab:pretraining_data} provides the summary statistics extracted from each dataset used for pretraining. Table~\ref{tab:downstream_data} provides the summary statistics from the data not used for pretraining, but used for downstream tasks. Table~\ref{tab:channels} lists the channels included in each modality from different datasets.

\begin{table*}[b]
    \centering
    \caption{The dataset statistics used for pretraining. Missing values result from study design or anonymized data.}
    \label{tab:pretraining_data}
    \resizebox{\textwidth}{!}{
    \small
    \renewcommand{\arraystretch}{1.5}
    \begin{tabular}{c c c c c c | c c c c c c | c c c c c}
        \toprule
        \multirow{2}{*}{\textbf{Dataset}} & \multirow{2}{*}{\textbf{Records}} & \multirow{2}{*}{\textbf{Subjects}} & \multirow{2}{*}{\textbf{Age (years)}} & \multirow{2}{*}{\textbf{BMI}} & \multirow{2}{*}{\begin{tabular}{@{}c@{}}\textbf{Sex \%} \\ \textbf{(Female/Male)}\end{tabular}} & \multicolumn{6}{c|}{\textbf{Stage count}} & \multicolumn{5}{c}{\textbf{Stage ratio (\%)}} \\
        & & & & & & W & N1 & N2 & N3 & REM & Total & W & N1 & N2 & N3 & REM \\
        \midrule
        PHY-Train & 993 & 993 & 55.2 $\pm$ 14.3 & N/A & 33 / 67 & 145,558 & 135,409 & 372,208 & 101,678 & 113,859 & 868,712 & 17 & 16 & 42 & 12 & 13 \\
        PHY-Test & 989 & 989 & 54.8 $\pm$ 14.3 & N/A & 37 / 63 & \multicolumn{6}{c|}{N/A} & \multicolumn{5}{c}{N/A} \\
        \midrule
        SHHS1 & 3,667 & 3,667 & 63.1 $\pm$ 11.5 & 28.2 $\pm$ 5.2 & 52 / 48 & 739,301 & 136,407 & 1,519,573 & 472,529 & 516,768 & 3,384,578 & 22 & 4 & 45 & 14 & 15 \\
        SHHS2 & 2,554 & 2,554 & 67.6 $\pm$ 10.4 & 28.3 $\pm$ 5.0 & 54 / 46 & 683,291 & 106,964 & 1,103,742 & 303,068 & 395,437 & 2,592,502 & 26 & 4 & 43 & 12 & 15 \\
        \midrule
        KISS & 6,064 & 6,064 & 44.8 $\pm$ 14.5 & 25.8 $\pm$ 4.3 & 20 / 80 & 981,362 & 665,497 & 1,556,469 & 601,227 & 660,186 & 4,464,741 & 22 & 15 & 35 & 13 & 15 \\
        KVSS & 881 & 881 & 51.4 $\pm$ 14.2 & 26.8 $\pm$ 4.4 & 24 / 76 & 143,041 & 129,225 & 271,190 & 47,538 & 97,667 & 688,661 & 21 & 19 & 39 & 7 & 14 \\
        \midrule
        MrOS1 & 2,768 & 2,768 & 76.4 $\pm$ 5.5 & 27.1 $\pm$ 3.8 & 0 / 100 & 945,515 & 129,239 & 1,236,682 & 221,334 & 381,577 & 2,914,347 & 32 & 4 & 43 & 8 & 13 \\
        MrOS2 & 994 & 994 & 81.0 $\pm$ 4.4 & 26.9 $\pm$ 3.8 & 0 / 100 & 382,152 & 80,312 & 425,515 & 45,040 & 129,942 & 1,062,961 & 36 & 8 & 40 & 4 & 12 \\
        \midrule
        MESA & 2,054 & 2,054 & 69.4 $\pm$ 9.1 & 28.7 $\pm$ 5.5 & 54 / 46 & 598,750 & 203,837 & 854,634 & 149,770 & 268,646 & 2,075,637 & 29 & 10 & 41 & 7 & 13 \\
        \bottomrule
    \end{tabular}
    }
\end{table*}

\begin{table*}[t]
    \centering
    \caption{The dataset statistics used for downstream evaluation.}
    \label{tab:downstream_data}
    \resizebox{\textwidth}{!}{
    \small
    \renewcommand{\arraystretch}{1.5}
    \begin{tabular}{c c c c c c | c c c c c c | c c c c c}
        \toprule
        \multirow{2}{*}{\textbf{Dataset}} & \multirow{2}{*}{\textbf{Records}} & \multirow{2}{*}{\textbf{Subjects}} & \multirow{2}{*}{\textbf{Age (years)}} & \multirow{2}{*}{\textbf{BMI}} & \multirow{2}{*}{\begin{tabular}{@{}c@{}}\textbf{Sex \%} \\ \textbf{(Female/Male)}\end{tabular}} & \multicolumn{6}{c|}{\textbf{Stage count}} & \multicolumn{5}{c}{\textbf{Stage ratio (\%)}} \\
        & & & & & & W & N1 & N2 & N3 & REM & Total & W & N1 & N2 & N3 & REM \\
        \midrule
        SHHS1-Val & 99 & 99 & 62.0 $\pm$ 11.7 & 27.8 $\pm$ 4.2 & 54 / 46 & 20,178 & 3,544 & 40,377 & 11,831 & 13,740 & 89,670 & 23 & 4 & 45 & 13 & 15 \\
        SHHS1-Test & 1,617 & 1,617 & 63.4 $\pm$ 11.5 & 28.0 $\pm$ 5.0 & 53 / 47 & 323,871 & 61,191 & 671,916 & 203,864 & 226,113 & 1,486,955 & 22 & 4 & 45 & 14 & 15 \\
        \midrule
        KISS-Val & 748 & 748 & 44.7 $\pm$ 14.1 & 26.0 $\pm$ 4.3 & 22 / 78 & 121,319 & 87,032 & 187,347 & 71,362 & 78,128 & 545,188 & 22 & 16 & 34 & 13 & 14 \\
        KISS-Test & 767 & 767 & 45.4 $\pm$ 14.3 & 26.0 $\pm$ 4.1 & 16 / 84 & 125,315 & 83,396 & 198,155 & 73,154 & 83,428 & 563,448 & 22 & 15 & 35 & 13 & 15 \\
        \midrule
        ISRUC-SG1 & 100 & 100 & 51.1 $\pm$ 15.9 & N/A & 44 / 56 & 20,979 & 11,513 & 28,287 & 17,480 & 11,928 & 90,187 & 23 & 13 & 31 & 19 & 13 \\
        ISRUC-SG2 & 16 & 8 & 46.9 $\pm$ 17.5 & N/A & 25 / 75 & 2,282 & 2,211 & 5,042 & 2,609 & 2,063 & 14,207 & 16 & 16 & 35 & 18 & 15 \\
        ISRUC-SG3 & 10 & 10 & 39.6 $\pm$ 9.6 & N/A & 10 / 90 & 1,817 & 1,248 & 2,678 & 2,035 & 1,111 & 8,889 & 20 & 14 & 30 & 23 & 12 \\
        \midrule
        \edit{CFS} & \edit{669} & \edit{669} & \edit{43.5} $\pm$ \edit{18.1} &
        \edit{33.1} $\pm$ \edit{9.3} & \edit{56 / 44} & \edit{165,072} & \edit{24,633} & \edit{284,676} & \edit{93,317} & \edit{93,119} & \edit{660,817} & \edit{25} & \edit{4} & \edit{43} & \edit{14} & \edit{14} \\
        \midrule
        \edit{SOF} & \edit{453} & \edit{453} & \edit{82.8} $\pm$ \edit{3.1} & \edit{27.7} $\pm$ \edit{4.7} & \edit{100 / 0} & \edit{133,091} & \edit{15,878} & \edit{176,092} & \edit{64,441} & \edit{58,300} & \edit{447,802} & \edit{30} & \edit{4} & \edit{39} & \edit{14} & \edit{13} \\
        \bottomrule
    \end{tabular}
    }
\end{table*}

\begin{table}[ht]
    \centering
    \caption{Channels and sampling rates included in different modalities across datasets. The manufacturer of PSG recording machine is also provided. When sampling rates are same across different channels in the modality, we only write once at the top row.}
    \label{tab:channels}
    \resizebox{\textwidth}{!}{%
    \begin{tabular}{lcccccccccccccccc}
        \toprule
        \multirow{2}{*}{\textbf{Modality}}  
        & \multicolumn{2}{c}{\textbf{KISS / KVSS}} 
        & \multicolumn{2}{c}{\textbf{SHHS1}} 
        & \multicolumn{2}{c}{\textbf{SHHS2}} 
        & \multicolumn{2}{c}{\textbf{PHY}} 
        & \multicolumn{2}{c}{\textbf{MESA}} 
        & \multicolumn{2}{c}{\textbf{MrOS1 / 2}} 
        & \multicolumn{2}{c}{\edit{\textbf{CFS}}}
        & \multicolumn{2}{c}{\edit{\textbf{SOF}}} \\ 
        \cmidrule(lr){2-3} 
        \cmidrule(lr){4-5} 
        \cmidrule(lr){6-7} 
        \cmidrule(lr){8-9} 
        \cmidrule(lr){10-11} 
        \cmidrule(lr){12-13}
        \cmidrule(lr){14-15}
        \cmidrule(lr){16-17}
         & Ch. & SR (Hz) 
         & Ch. & SR (Hz) 
         & Ch. & SR (Hz) 
         & Ch. & SR (Hz) 
         & Ch. & SR (Hz) 
         & Ch. & SR (Hz) 
         & \edit{Ch.} & \edit{SR (Hz)}
         & \edit{Ch.} & \edit{SR (Hz)} \\ 
         \midrule

        \multirow{6}{*}{EEG}  
          & C3-M2 & 200 
          & C3-A2 & 125 
          & C3-A2 & 125 
          & C3-M2 & 200 
          & C4-M1 & 256 
          & C3-M2 & 256 
          & \edit{C3-A2} & \edit{128}
          & \edit{C3-A2} & \edit{128} \\
          & C4-M1 &  
          & C4-A1 &  
          & C4-A1 &  
          & C4-M1 &  
          & Oz-Cz &  
          & C4-M1 &  
          & \edit{C4-A1} & 
          & \edit{C4-A1} &  \\
          & O1-M2 &  
          &  &  
          &  &  
          & O1-M2 &  
          & Fz-Cz &  
          & O1-M2 &  
          &  & 
          &  &  \\
          & O2-M1 &  
          &  &  
          &  &  
          & O2-M1 &  
          &  &  
          & O2-M1 &  
          &  & 
          &  &  \\
          &  &  
          &  &  
          &  &  
          & F3-M2 &  
          &  &  
          &  &  
          &  & 
          &  &  \\
          &  &  
          &  &  
          &  &  
          & F4-M1 &  
          &  &  
          &  &  
          &  & 
          &  &  \\ 
          \midrule

        \multirow{2}{*}{EOG}  
          & E1-M2 & 200 
          & EOG (L) & 125 
          & EOG (L) & 125  
          & E1-M2 & 200  
          & EOG (L) & 256 
          & EOG (L) & 256 
          & \edit{LOC} & \edit{128}
          & \edit{LOC} & \edit{128} \\
          & E2-M1 &  
          & EOG (R) &  
          & EOG (R) &  
          &  &  
          & EOG (R) &  
          & EOG (R) &  
          & \edit{ROC} & 
          & \edit{ROC} &  \\ 
          \midrule

        EMG  
          & Chin EMG & 200 
          & EMG & 125 
          & EMG & 125 
          & Chin1-Chin2 & 200 
          & Chin EMG & 256 
          & EMG (L)-EMG (R) & 256 
          & \edit{EMG1-EMG2} & \edit{256}
          & \edit{LChin-RChin} & \edit{128} \\ 
          \midrule

        ECG  
          & ECG & 200 
          & ECG & 125 
          & ECG & 125 
          & ECG & 250 
          & ECG & 256 
          & ECG1-ECG2 & 512 
          & \edit{ECG1-ECG2} & \edit{256/512}
          & \edit{ECG1-ECG2} & \edit{128/256} \\ 
          \midrule

        \multirow{4}{*}{Respiratory} 
          & Flow & 200 
          & Flow & 10 
          & Flow & 10 
          & Flow & 200 
          & Flow & 32 
          & Flow & 64 
          & \edit{Nasal Pres} & \edit{64/128}
          & \edit{Nasal Pres} & \edit{16/64} \\
          & Thermistor &  
          &  &  
          & Thorax &  
          & Thorax &  
          & Thorax &  
          & Thermistor & 16 
          & \edit{Airflow} & \edit{32}
          & \edit{Airflow} &  \\
          & Thorax &  
          &  &  
          & Abdomen &  
          & Abdomen &  
          & Abdomen &  
          & Thorax &  
          & \edit{Thor Effort} & \edit{32}
          & \edit{Thorax} &  \\
          & Abdomen &  
          &  &  
          &  &  
          &  &  
          & Press &  
          & Abdomen &  
          & \edit{Abdo Effort} & \edit{32}
          & \edit{Abdomen} &  \\ 
          \midrule

        Oxygen Sat. 
          & Saturation & 200 
          & Oximetry & 1 
          & Oximetry & 1 
          & SaO2 & 200 
          & SpO2 & 1 
          & SpO2 & 1 
          & \edit{SpO2} & \edit{1}
          & \edit{SpO2} & \edit{1} \\
        \midrule

        Manufacturer 
          & \multicolumn{2}{c}{Nox, Embla} 
          & \multicolumn{2}{c}{Compumedics} 
          & \multicolumn{2}{c}{Compumedics} 
          & \multicolumn{2}{c}{Unknown} 
          & \multicolumn{2}{c}{Compumedics} 
          & \multicolumn{2}{c}{Compumedics} 
          & \multicolumn{2}{c}{\edit{Compumedics}}
          & \multicolumn{2}{c}{\edit{Compumedics}} \\ 
        \bottomrule
    \end{tabular}%
    }
\end{table}

\subsection{Physionet 2018 (PHY) Dataset}
The PhysioNet 2018 dataset, originally curated for the 2018 PhysioNet/CinC Challenge~\cite{goldberger2000physiobank, ghassemi2018you}, was provided by the Computational Clinical Neurophysiology Laboratory and the Clinical Data Animation Laboratory at Massachusetts General Hospital. Although the challenge primarily focused on arousal detection, the dataset includes expert-labeled sleep stages for 994 subjects. An additional 991 recordings were reserved for testing purposes; however, their labels remain private. In this study, we utilized both the labeled training set and the unlabeled test set for pretraining \proposal. All signals were sampled at 200 Hz, with sleep stages scored according to the American Academy of Sleep Medicine (AASM) guidelines.

\subsection{Sleep Heart Health Study (SHHS) Dataset}
The Sleep Heart Health Study (SHHS)~\cite{zhang2018national, quan1997sleep} is a multicenter cohort initiative organized by the National Heart, Lung, and Blood Institute. This study consists of data collected over two visits. SHHS1 was collected from the initial visit, conducted between 1995 and 1998, which involved 6,441 men and women aged 40 and older. SHHS2 was collected from the second visit, conducted between 2001 and 2003, which involved 3,295 participants.  Polysomnography (PSG) was recorded in-home by trained technicians and included various physiological signals: EEG (C3-A2, C4-A1), dual-channel EOG, EMG, respiratory effort, airflow, oxygen saturation, ECG, and body position. In alignment with the experimental protocol of XSleepNet~\cite{phan2021xsleepnet}, we partitioned the dataset by reserving 30\% for testing. From the remaining 70\%, 100 subjects were set aside for validation, with the balance used for model training. The train split of SHHS1 is used for pretraining and fine-tuning for downstream tasks. The entire SHHS2 dataset is used for pretraining and not used for downstream tasks.

\subsection{Korea Image-based Sleep Study (KISS) Dataset}
The Korea Image-based Sleep Study (KISS) dataset~\cite{jeong2023standardized} is a standardized, image-based polysomnography (PSG) repository. Collected between 2013 and 2020 across four sleep centers, the dataset utilizes recordings from Embla and NOX-A1 PSG systems, totaling 10,253 records. Expert scoring was conducted in accordance with AASM version 2.6 guidelines~\cite{berry2012aasm, berry2017aasm}. Each record captures 21 distinct biosignals including various EEG, EOG, and EMG channels alongside respiratory and movement data. The data is publicly accessible via AI Hub~\cite{aihub}. Following the experimental setup by Jeong \textit{et al.}~\cite{jeong2023standardized}, we selected 7,579 records and implemented an 80\%/20\% split for training and validation/test on a patient-wise basis. The train split of KISS is used for pretraining and fine-tuning for downstream tasks.

\subsection{Korea Video Sleep Study (KVSS) Dataset}
The Korea Video Sleep Study (KVSS) dataset is a retrospectively constructed, multi-center clinical cohort that provides synchronized infrared sleep video and polysomnography (PSG) with expert annotations~\cite{choi2024kvss}. Data were collected under IRB-approved protocols from three hospitals (Chungnam National University Hospital, The Catholic University of Korea St.~Vincent's Hospital, and Hallym University Hospital). 
The full KVSS cohort consists of 936 PSG examinations with synchronized infrared video. This study used a subset of 881 examinations selected after screening and quality control. 
Recordings were obtained during routine clinical studies for various sleep-related indications, including suspected OSA, insomnia, PLMD, and RBD, with infrared videos recorded in parallel with PSG to ensure temporal alignment. Despite minor site-specific differences in camera placement and illumination, all videos were standardized to MP4 at $640\times480$ resolution and 5~fps. PSG was stored in European Data Format (EDF), and sleep stages (Wake, N1, N2, N3, REM) and AASM-defined events were annotated by certified technologists and reviewed by sleep physicians. The dataset underwent de-identification and expert cross-checking to verify synchronization and address obvious scoring inconsistencies.
In this work, KVSS is used only as a PSG dataset, and we leverage the recorded PSG signals and associated subject metadata without using expert annotations or the infrared videos.

\subsection{Multi-Ethnic Study of Atherosclerosis (MESA) Dataset}
The Multi-Ethnic Study of Atherosclerosis (MESA)~\cite{chen2015racial} is a multicenter cohort of 6,814 adults aged 45–84 years from four racial/ethnic groups (White, Black, Hispanic, and Chinese-American). As part of Exam 5 (2010–2013), the MESA Sleep exam enrolled 2,237 participants who completed single-night unattended in-home polysomnography (PSG) and wrist actigraphy. PSG was set up during an in-home evening visit by trained staff, and sleep staging and respiratory events were scored at a centralized sleep reading center using standardized procedures.
The National Sleep Research Resource (NSRR) release provides PSG recordings in European Data Format (EDF) and XML annotation files for sleep staging and respiratory event scoring. Respiratory event annotations were harmonized via rule-based post-processing of the original labels to ensure consistent criteria across datasets, and details are described in ~\cite{ahn2025refining}. The PSG includes EEG (Fz-Cz, Cz-Oz, C4-A1), EOG, EMG, ECG, nasal airflow, thoracic and abdominal respiratory effort, oxygen saturation, and body position. In this work, we utilize the C4-A1 EEG. Recordings missing any required channel were excluded. 2,054 participants were included in the final analytic cohort.

\subsection{Osteoporotic Fractures in Men Study (MrOS) Dataset}
The Osteoporotic Fractures in Men Sleep Study (MrOS Sleep)~\cite{blackwell2011associations} is a multicenter sleep cohort of older men (aged 65 years or older) that includes unattended in-home polysomnography (PSG), which was set up by trained technicians and annotated using standardized scoring procedures. The study was conducted between December 2003 and March 2005, during which 3,135 participants from the parent MrOS cohort of 5,994 men completed overnight unattended in-home PSG.
The dataset available through the National Sleep Research Resource (NSRR) includes PSG recordings in European Data Format (EDF) and XML annotation files with multimodal biosignals, including EEG, EOG, EMG, ECG, nasal cannula (airflow), thoracic and abdominal respiratory effort, oxygen saturation, and body position. Respiratory event annotations were harmonized via rule-based post-processing of the original labels to ensure consistent criteria across datasets, and details are described in ~\cite{ahn2025refining}. For the present study, we analyzed data from two sleep visits (Visit 1, 2003–2005; Visit 2, 2009–2012). We excluded recordings missing any of the required signals (EEG, EOG, EMG, ECG, airflow, thoracic/abdominal effort, or oxygen saturation). Of the 2,911 participants with successful PSG recordings, 2,678 (Visit 1) and 998 (Visit 2) were included in the final analytic cohort after quality control.

\subsection{Institute of Systems and Robotics, University of Coimbra (ISRUC) Dataset}
The Institute of Systems and Robotics, University of Coimbra (ISRUC) dataset~\cite{khalighi2016isruc} is a publicly available repository provided by the Sleep Medicine Center of the Hospital of Coimbra University (CHUC). 
The dataset is divided into three subgroups: SG1 and SG2, which feature patients with sleep disorders, and SG3, which consists of healthy control subjects. Each recording comprises 19 signals, including six EEG channels (F3-A2, C3-A2, O1-A2, F4-A1, C4-A1, O2-A1), dual-channel EOG and EMG, and various respiratory and cardiac sensors. 
Considering its relatively small size, this dataset is used to evaluate the adaptation efficiency of our model.

\subsection{\edit{Cleveland Family Study (CFS) Dataset}}

\edit{The Cleveland Family Study (CFS) \citep{zhang2018national, redline1995familial} is a family-based cohort study designed to investigate the familial aggregation and risk factors of sleep apnea. The cohort consists of 2,284 individuals from 361 families who were followed longitudinally over multiple visits across a 16-year period. The NSRR release provides overnight polysomnography (PSG) recordings, including full-night PSG data from Visit 5. In this work, CFS is used as an external held-out downstream evaluation dataset and is not included in the pretraining corpus. To account for the distinct clinical scoring guidelines for children, subjects under the age of 14 are excluded from our evaluation.  We use the available PSG signals and corresponding sleep-stage and respiratory-event annotations for sleep stage classification and sleep-disordered breathing segmentation, after applying the same preprocessing pipeline used for the other NSRR datasets.}

\subsection{\edit{Study of Osteoporotic Fractures (SOF) Dataset}}

\edit{The Study of Osteoporotic Fractures (SOF) \citep{zhang2018national, spira2008sleep} is a prospective, multi-center cohort originally established to study risk factors for osteoporotic fractures and falls in older women. The sleep study includes women aged 65--89 years and provides raw EDF PSG recordings together with PSG-derived variables through the NSRR. Sleep studies were completed on 461 SOF participants at Visit 8, and recordings passing our quality-control criteria were used for downstream evaluation. In this work, SOF is used exclusively as an external held-out downstream evaluation dataset and is not included in the pretraining corpus. We use the available PSG recordings and annotations for sleep stage classification and sleep-disordered breathing segmentation, enabling evaluation of cross-cohort generalization to an older, all-female population.}


\subsection{Demographics analysis}

\begin{figure}[h]
    \centering
    \begin{subfigure}[b]{0.35\textwidth}
        \centering
        \includegraphics[width=\textwidth, bb=0 0 216 252]{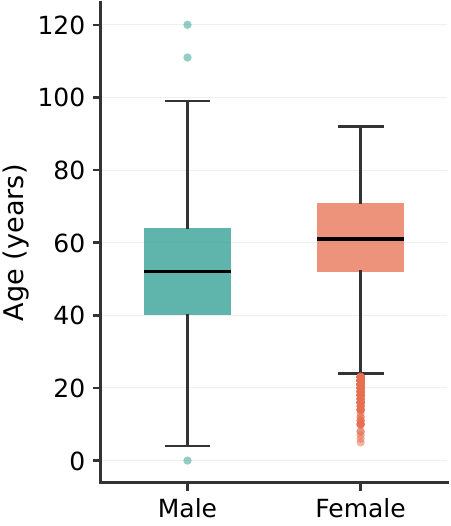}
        \caption{Age distribution by sex.}
        \label{fig:age_distribution}
    \end{subfigure}
    \hspace{1cm}
    \begin{subfigure}[b]{0.35\textwidth}
        \centering
        \includegraphics[width=\textwidth, bb=0 0 216 252]{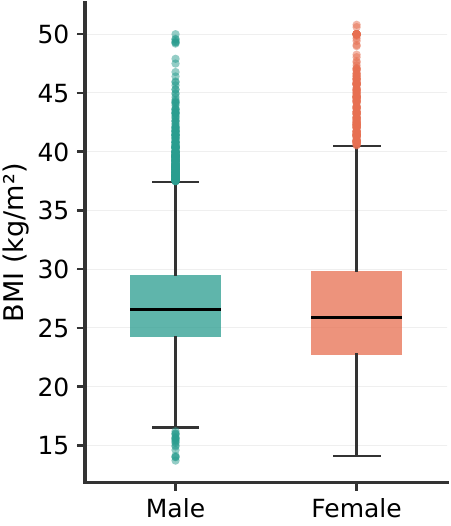}
        \caption{BMI distribution by sex.}
        \label{fig:bmi_distribution}
    \end{subfigure}
    \caption{Demographic distributions of the pre-training dataset. (a) Age distribution shows a median of approximately 52 years for males and 61 years for females. (b) BMI distribution shows similar medians across sexes (approximately 27 kg/m² for males and 26 kg/m² for females), with notable outliers in both groups.}
    \label{fig:demographic_distributions}
\end{figure}

Figure~\ref{fig:demographic_distributions} shows the age and BMI distribution stratified by sex based on our pretraining datasets.

\section{Model architecture details}
Table~\ref{tab:architecture_details} summarizes the specification of the Micro-Encoder and the Macro-Encoder.

\begin{table}[ht]
    \centering
    \caption{Architecture Specification for the Micro- and Macro- Encoders. Norm: Normalization layer (RMSNorm), Exp: Expansion convolution, DW: Depthwise convolution, PW: Pointwise convolution}
    \label{tab:architecture_details}
    \resizebox{0.8\textwidth}{!}{%
    \begin{tabular}{lllcclccl}
    \toprule
    \textbf{Encoder} & \textbf{Module} & \textbf{Item} & \textbf{Dim.} & \textbf{Kernel} & \textbf{Stride} & \textbf{Depth} & \textbf{Heads} & \textbf{Notes} \\ \midrule
    \multirow{10}{*}{Micro} & Patch Emb. & Patch Emb. (Conv blocks) & [32,64,128] & [10,5,1] & [10,5,1] & -- & -- & Conv-Norm-GELU \\ \cmidrule(l){2-9} 
     & \multirow{3}{*}{Private} & Transformer blocks (Lower) & 128 & -- & -- & 2 & 8 &  \\
     &  & Patch Merge & 384 & 10 & 5 & -- & -- & Exp-DW-Norm-PW-GELU-PW \\
     &  & Transformer blocks (Higher) & 384 & -- & -- & 2 & 8 &  \\ \cmidrule(l){2-9} 
     & Shared & MoE Transformer blocks & 384 & -- & -- & 4 & 8 & 4 Experts, 2 Activated \\ \cmidrule(l){2-9} 
     & \multirow{2}{*}{Fusion} & Transformer blocks & 384 & -- & -- & 4 & 8 & Cross-Attention \\
     &  & Upsample & 128 & -- & -- & -- & -- & \\ \cmidrule(l){2-9} 
     & \multirow{3}{*}{Decoder} & Linear embedding & 64 & -- & -- & -- & -- & \\
     &  & Transformer blocks & 64 & -- & -- & 2 & 4 &  \\
     &  & Projection (Linear) & 50 & -- & -- & -- & -- & Same as the patch size \\ \midrule
    \multirow{3}{*}{Macro} & \multirow{3}{*}{Macro} & Linear projection & 512 & -- & -- & -- & -- &  \\
     &  & Mamba blocks (Forward) & 512 & -- & -- & 2 & -- & PreNorm-Mamba-PostNorm \\
     &  & Mamba blocks (Backward) & 512 & -- & -- & 2 & -- & PreNorm-Mamba-PostNorm \\ \bottomrule
    \end{tabular}%
    }
\end{table}

\section{Training details} \label{sec:training_details}
Pretraining is done in two-step process. The Micro-Encoder is trained in the first step and the Macro-Encoder is trained in the second step.  Table~\ref{tab:hyperparameters} summarizes pretraining details for both Micro- and Macro-Encoder. When pretraining Macro-Encoder, we did not include PHY dataset because the dataset does not provide BMI information. Moreover, in other datasets, when the demographic attributes are not in proper format or unavailable, we exclude those records. 

Table~\ref{tab:downstream} summarizes the fine-tuning settings for downstream tasks used for Section~\ref{sec:Experiments}.

When calculating the reconstruction loss for Micro-Encoder ($\mathcal{L}_{Mi}^{recon}$), the raw signal is smoothed using moving average of eleven adjacent points. This effectively eliminates noises and helps the model focus on more meaningful signals.

\begin{table}[ht]
    \centering
    \caption{Hyperparameter settings for pretraining for Micro and Macro-Encoders}
    \label{tab:hyperparameters}
    \resizebox{0.75\linewidth}{!}{%
    \begin{tabular}{@{}llll@{}}
        \toprule
        \textbf{Encoder} & \textbf{Item} & \textbf{Value} & \textbf{Notes} \\ \midrule
        \multirow{17}{*}{Micro-Encoder} & Mask ratio & 50\% &  \\
         & Batch size & 512 &  \\
         & Input length & 60 seconds & Equivalent to 120 \edit{patches} \\
         & Optimizer & AdamW &  \\
         & $\beta_1$ & 0.9 &  \\
         & $\beta_2$ & 0.99 &  \\
         & Weight decay ($\lambda$) & 0.05 &  \\
         & Initial learning rate & $5.00 \times 10^{-4}$ &  \\
         & Learning rate schedule & Cosine annealing &  \\
         & Final learning rate & $1.00 \times 10^{-8}$ &  \\
         & Training epochs & 3 &  \\
         & Patch size & 500 ms & Equivalent to 50 input points \\
         & Temperature for contrastive loss ($\tau$) & 0.07 &  \\
         & Sequence length for contrastive loss & 30 seconds &  \\ 
         & Weight for contrastive loss ($\lambda_{CL}$) & 0.1 & \\
         & Weight for KoLeo loss ($\lambda_{KoLeo}$) & 0.01 & \\ 
         & Timespan for contrastive loss & 30 seconds & Equivalent to 60 epochs \\\midrule
        
        \multirow{12}{*}{Macro-Encoder} & Batch size & 40 & Subjects \\
         & Maximum number of epochs per subject & 1,080 & Equivalent to 540 minutes \\
         & Optimizer & AdamW &  \\
         & $\beta_1$ & 0.9 &  \\
         & $\beta_2$ & 0.99 &  \\
         & Weight decay ($\lambda$) & 0.05 & Not applied to SSM parameters \\
         & Initial learning rate & $1.00 \times 10^{-4}$ &  \\
         & Learning rate schedule & Cosine annealing with warmup &  \\
         & Final learning rate & $1.00 \times 10^{-8}$ &  \\
         & Training epochs & 4 &  \\
         & Temperature for contrastive loss ($\rho$) & 0.1 &  \\
         & Temperature for weight calculation ($\upsilon$) & 0.5 &  \\
         & Cycle length & 90 minutes & 180 epochs  \\ 
         & Demographic distance for sex difference ($\lambda_{sex}$) & 1 &   \\ \bottomrule
    \end{tabular}
    }
\end{table}

\begin{table}[ht]
    \centering
    \caption{Training details used for downstream tasks.}
    \label{tab:downstream}
    \resizebox{0.80\linewidth}{!}{%
    \begin{tabular}{@{}llll@{}}
    \toprule
        \textbf{Downstream tasks} & \textbf{Item} & \textbf{Value} & \textbf{Notes} \\ \midrule
        \multirow{3}{*}{Sleep staging} & Loss & Weighted Cross Entropy & $w_k=\log_5(N/N_k)$ for class $k$ \\
         & Initial learning rate & $1.00 \times 10^{-2}$ &  \\         
         & Batch size & 4 & Subjects \\   \hline      
         
         \multirow{3}{*}{Apnea segmentation} & Loss & Weighted Cross Entropy & $w_k=1+\log_2 (N/N_k)$ for class $k$ \\
         & Initial learning rate & $4.00 \times 10^{-4}$ &  \\         
         & Batch size & 1024 & Epochs \\ \hline
         
         \multirow{3}{*}{Disease prediction} & Loss  & Cox PH &  \\         
         & Initial learning rate & $1.00 \times 10^{-2}$ &  \\
         & Batch size & 4 & Subjects \\ \hline

         \multirow{3}{*}{Age / AHI estimation} & Loss  & MAE &  \\         
         & Initial learning rate & $1.00 \times 10^{-2}$ &  \\
         & Batch size & 4 & Subjects \\ \hline

         \multirow{3}{*}{Sex classification} & Loss  & Cross Entropy &  \\         
         & Initial learning rate & $1.00 \times 10^{-2}$ &  \\
         & Batch size & 4 & Subjects \\ \hline

         \multirow{7}{*}{Common settings} & Optimizer & AdamW &  \\
         & $\beta_1$ & 0.9 &  \\
         & $\beta_2$ & 0.99 &  \\
         & Weight decay ($\lambda$) & $1 \times 10^{-4}$ &  \\
         & Learning rate schedule & Cosine annealing &  \\
         & Final learning rate & $1.00 \times 10^{-8}$ &  \\
         & Training epochs & 3 &  \\ \bottomrule
    \end{tabular}
    }
\end{table}

\clearpage
\section{Sleep macro-structure analysis} \label{sec:more_macro_analysis}

\begin{figure}[h]
    \centering
    \begin{subfigure}[b]{0.45\textwidth}
        \centering
        \includegraphics[width=\textwidth, bb=0 0 701 705]{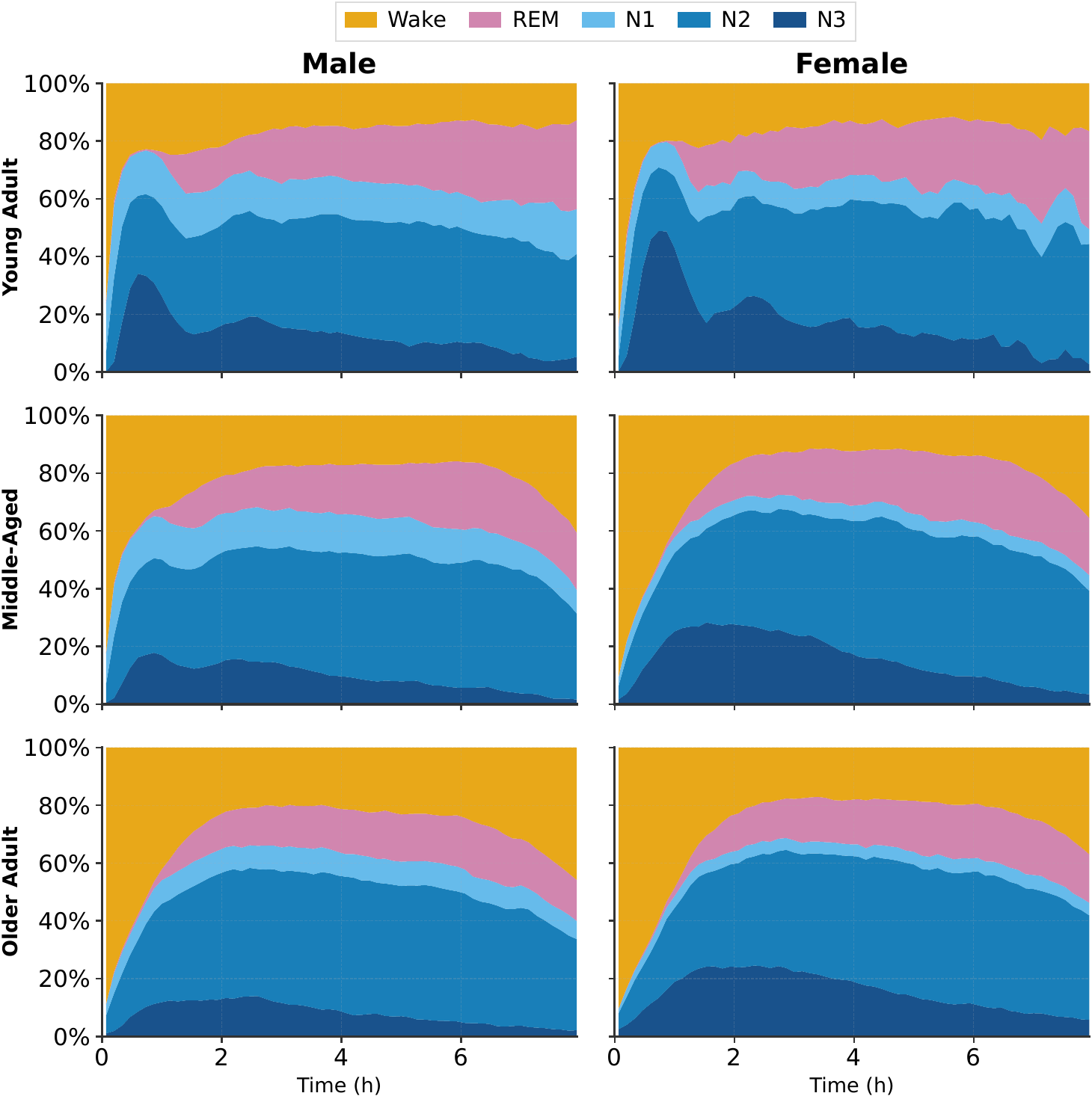}
        \caption{By age group}
        \label{fig:sleep_by_age}
    \end{subfigure}
    \hspace{1cm}
    \begin{subfigure}[b]{0.45\textwidth}
        \centering
        \includegraphics[width=\textwidth, bb=0 0 701 705]{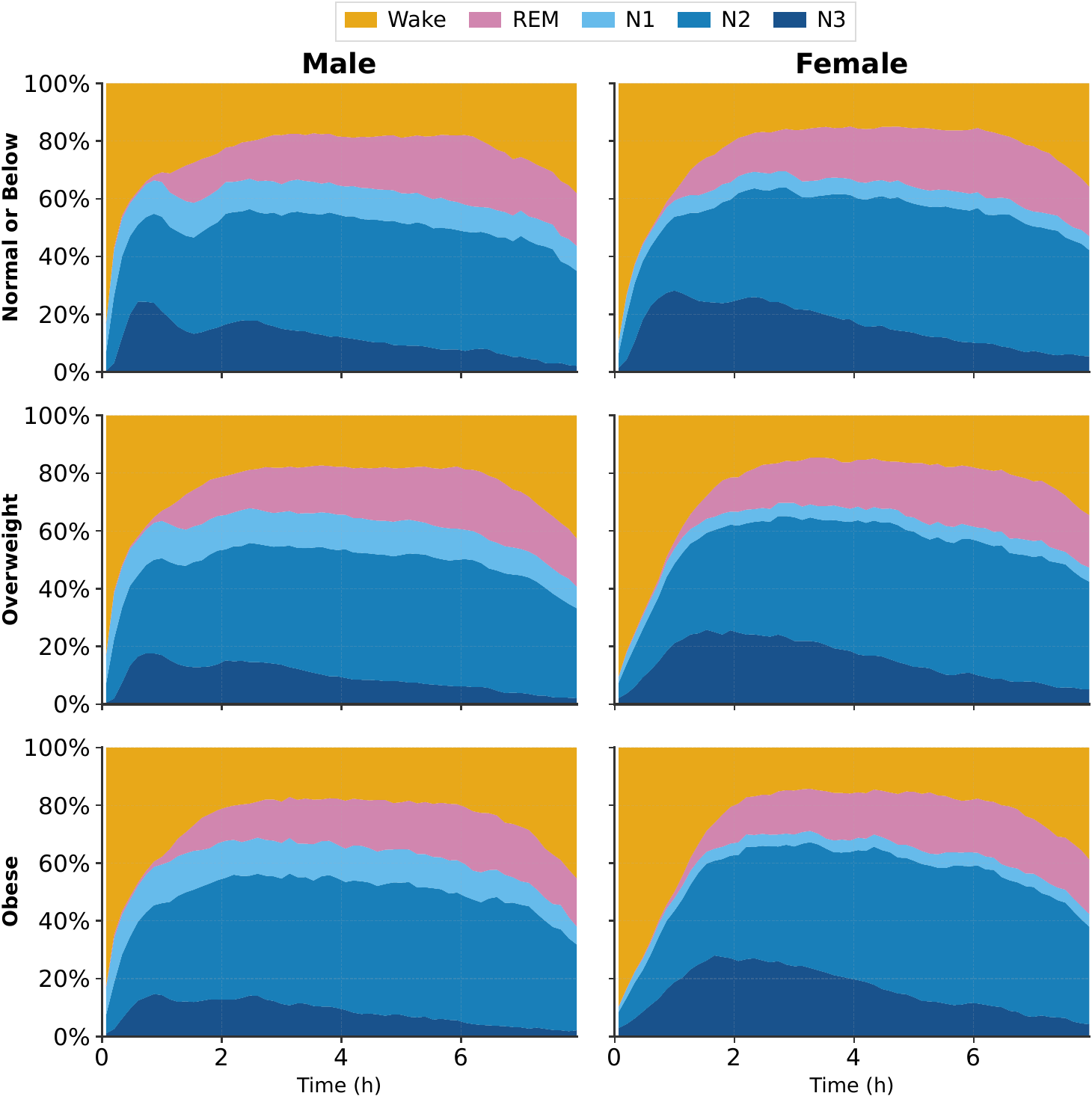}
        \caption{By BMI category}
        \label{fig:sleep_by_bmi}
    \end{subfigure}
    
    \vspace{0.2cm}
    
    \begin{subfigure}[b]{0.45\textwidth}
        \centering
        \includegraphics[width=\textwidth, bb=0 0 700 507]{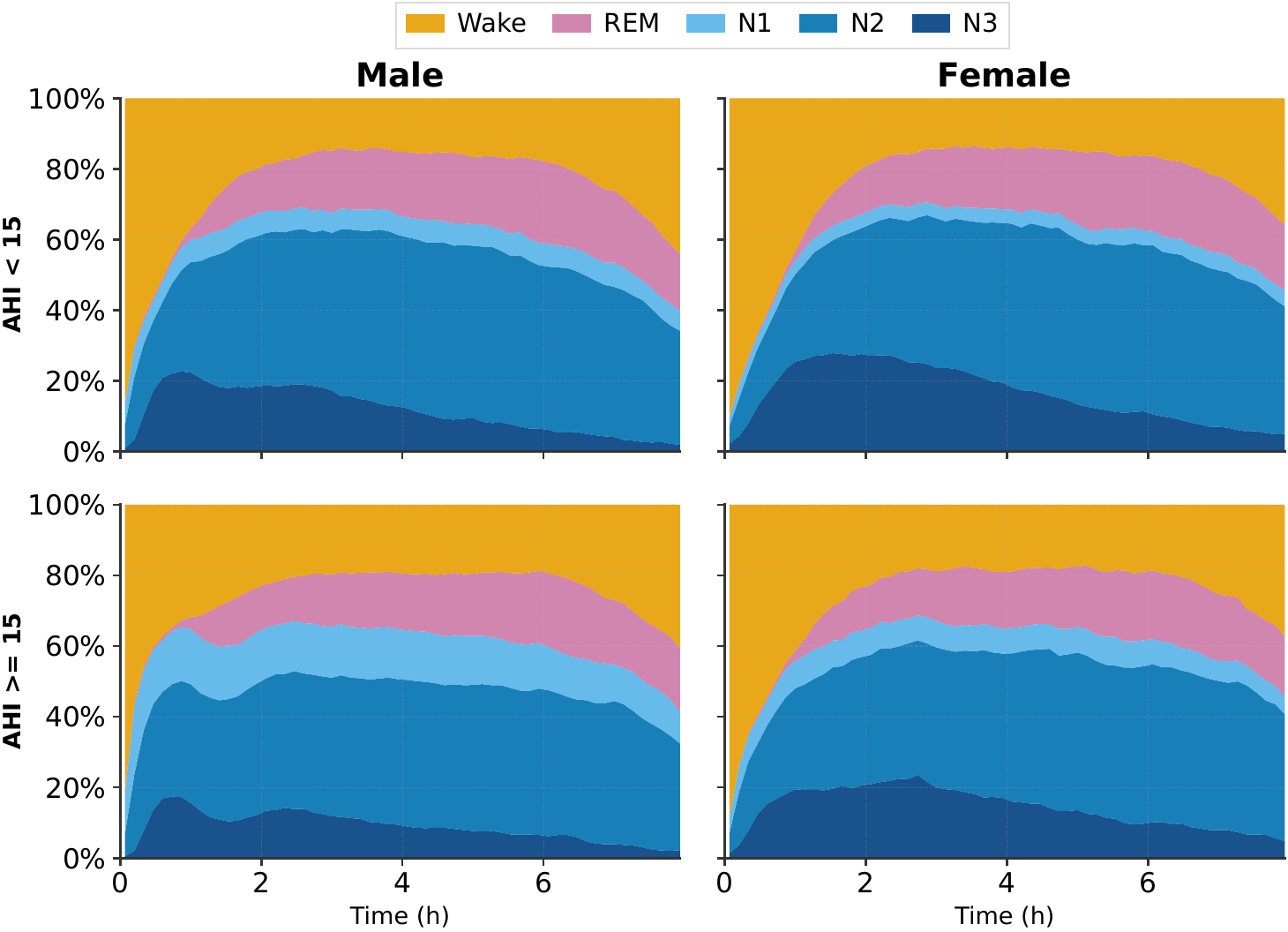}
        \caption{By sleep apnea severity}
        \label{fig:sleep_by_ahi}
    \end{subfigure}
    
    \caption{Sleep macrostructure variations across demographic and clinical groups. (a) Age groups: Young Adult (18--44 years), Middle-Aged (45--64 years), Older Adult ($\geq$ 65 years). (b) BMI categories: Normal or Below ($<$ 25 kg/m$^2$), Overweight (25--30 kg/m$^2$), Obese ($\geq$ 30 kg/m$^2$). (c) Sleep apnea severity based on Apnea-Hypopnea Index (AHI $\geq$ 15 events/h indicates moderate-to-severe). These demographic-dependent patterns motivate our Demographic-Guided Contrastive Learning objective.}
    \label{fig:sleep_macrostructure_appendix}
\end{figure}

\begin{figure*}[ht]
    \centering
    \begin{subfigure}[b]{0.65\textwidth}
        \centering
        \includegraphics[width=\textwidth, bb=0 0 974 648]{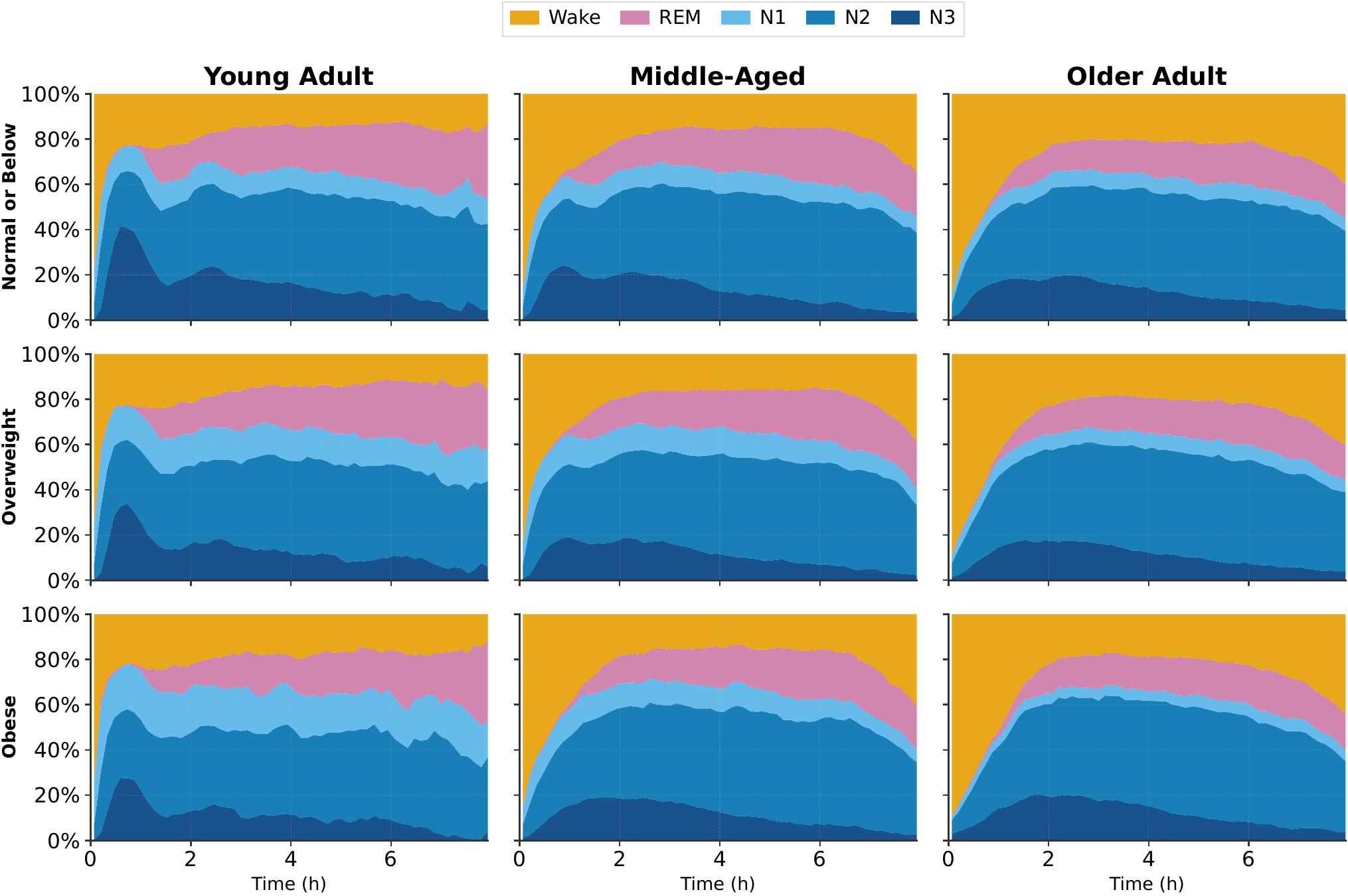}
        \caption{By BMI and age group}
        \label{fig:sleep_stages_bmi_age}
    \end{subfigure}

    \vspace{0.2cm}
    
    \begin{subfigure}[b]{0.45\textwidth}
        \centering
        \includegraphics[width=\textwidth, bb=0 0 672 648]{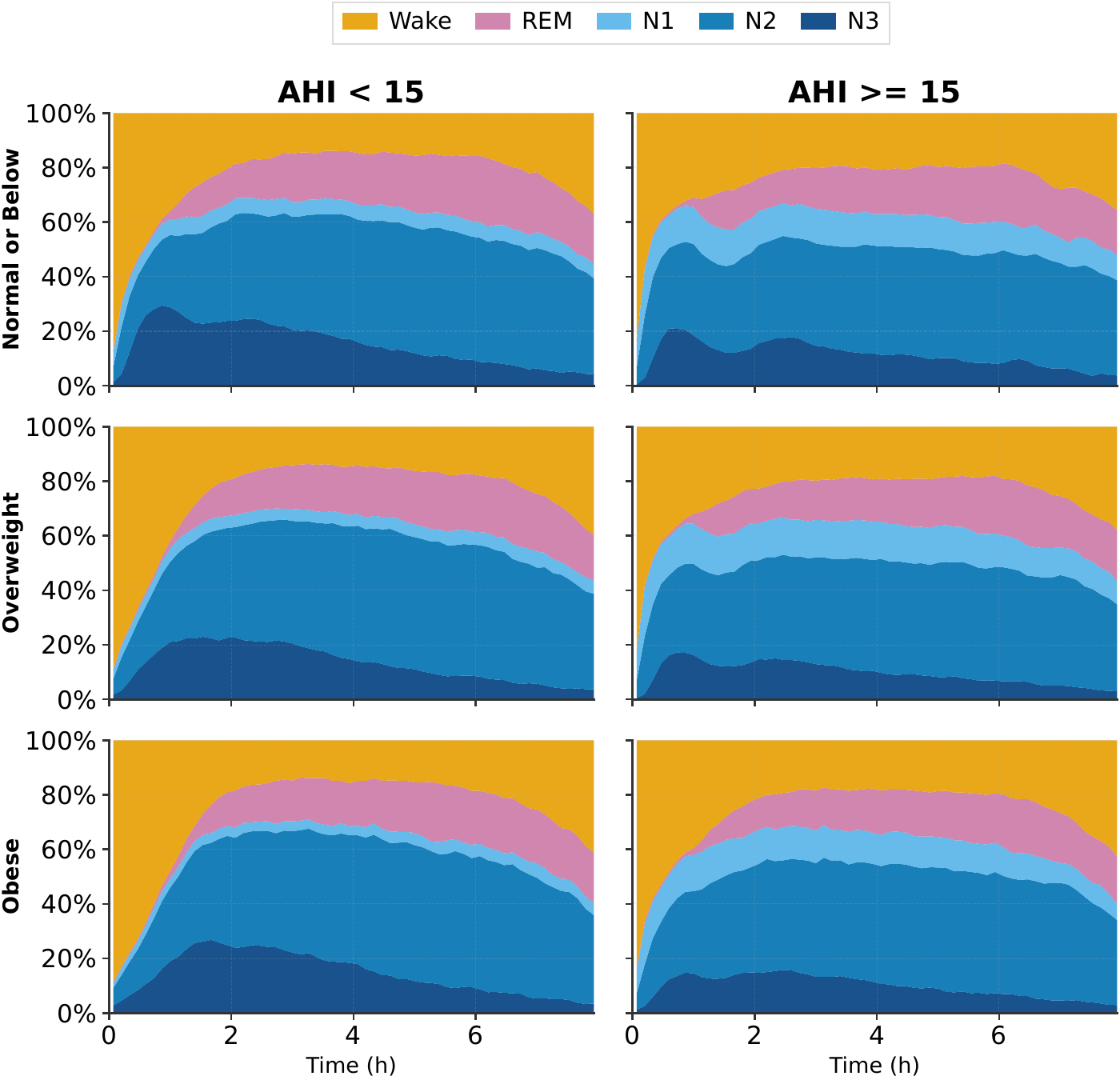}
        \caption{By BMI and AHI}
        \label{fig:sleep_stages_bmi_ahi}
    \end{subfigure}
    \hspace{1cm}
    \begin{subfigure}[b]{0.45\textwidth}
        \centering
        \includegraphics[width=\textwidth, bb=0 0 672 648]{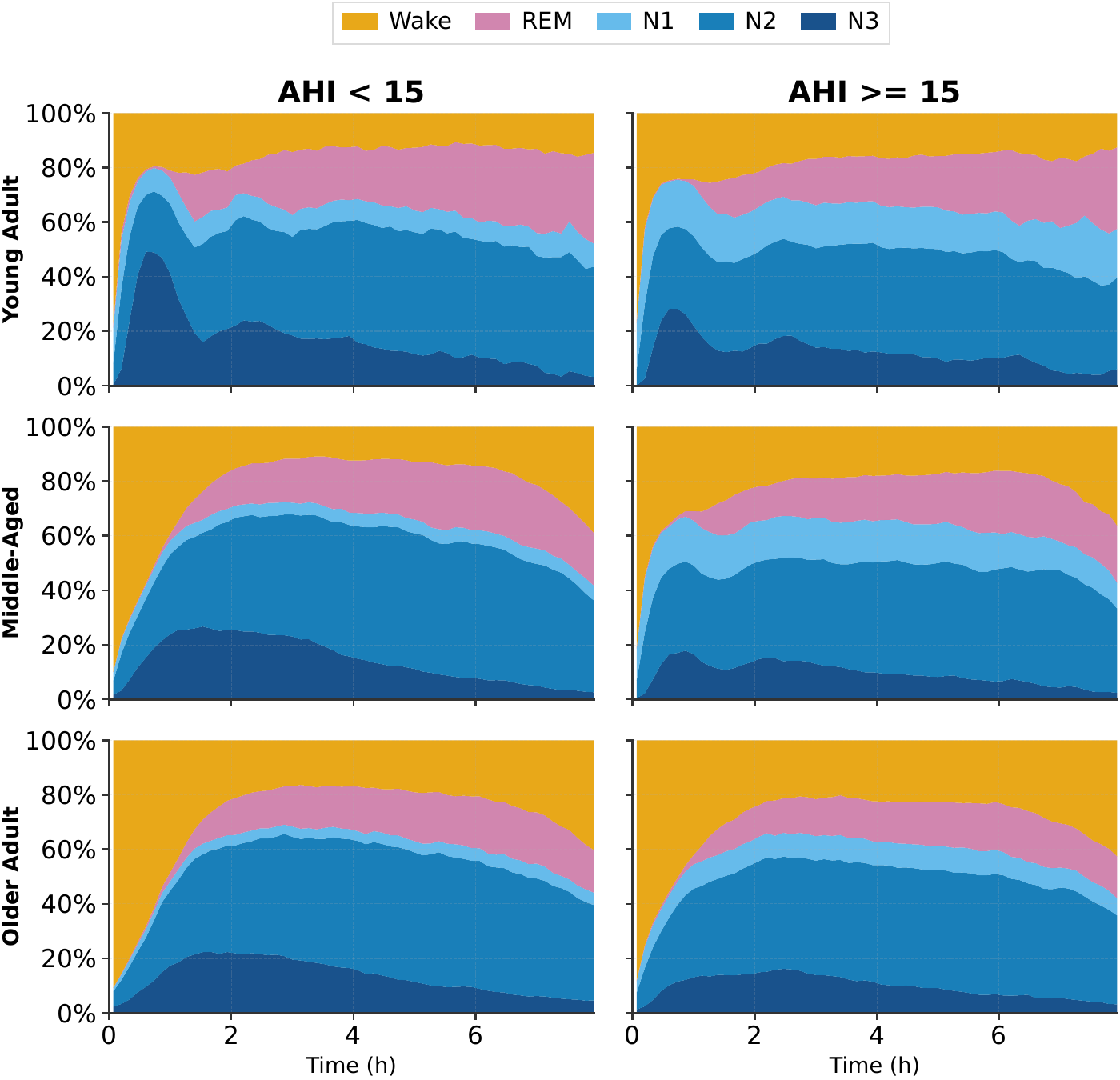}
        \caption{By age group and AHI}
        \label{fig:sleep_stages_age_ahi}
    \end{subfigure}
    \caption{Sleep macro-structure variations across combined demographic and clinical factors. (a) BMI categories (Normal or Below: $<$ 25 kg/m$^2$, Overweight: 25--30 kg/m$^2$, Obese: $\geq$ 30 kg/m$^2$) and age groups (Young Adult: 18--44 years, Middle-Aged: 45--64 years, Older Adult: $\geq$ 65 years). (b) BMI categories and sleep apnea severity (AHI $<$ 15 vs.\ AHI $\geq$ 15 events/h). (c) Age groups and sleep apnea severity. These combined factors jointly influence sleep stage distributions throughout the night.}
    \label{fig:sleep_stages_interactions}
\end{figure*}

To further investigate the demographic and clinical factors that influence sleep stage distributions, we present extended visualizations across various subgroups. Figure~\ref{fig:sleep_macrostructure_appendix} shows sleep stage distributions by sex combined with age group, BMI category, and sleep apnea severity (AHI), respectively. Figure~\ref{fig:sleep_stages_interactions} presents the interactions between non-sex factors, revealing how multiple variables jointly influence sleep stage distributions throughout the night.

\clearpage
\section{\edit{Demographic biases of DGCL}}
\edit{
A potential concern regarding the DGCL objective is that the model might encode demographic biases present in the training data. 
However, demographic attributes in DGCL function strictly as a soft supervision signal and are never provided as inputs to the Macro-Encoder, which processes epoch-level embeddings exclusively. 
While the demographic distance function guides the structural organization of the representation space, the representations themselves are derived entirely from physiological signals. 
Furthermore, the use of a soft-target contrastive loss preserves representation flexibility. 
To quantitatively evaluate any bias risks, we conduct a demographic subgroup analysis, comparing the disease prediction performance (Avg. C-Index, SHHS1) of \proposal against a Micro-only baseline. 
As shown in Table~\ref{tab:demographic_bias_analysis}, \proposal delivers consistent performance across all subgroups, confirming that DGCL does not introduce or amplify demographic bias. Although these results demonstrate robustness, exploring more granular de-biasing techniques remains a valuable direction for future clinical optimization.
}

\begin{table}[h]
\centering
\caption{Subgroup Analysis and Performance Gaps for Disease Prediction (Avg. C-Index, SHHS1)}
\label{tab:demographic_bias_analysis}
\begin{tabular}{lcc}
\toprule
\textbf{Group} & \textbf{\proposal} & \textbf{Micro-only} \\ 
\midrule
\textit{Age}   &                    &                     \\
Younger        & 0.668              & 0.554               \\
Older          & 0.674              & 0.676               \\
\hline
\textbf{Gap}   & -0.005             & -0.122              \\ 
\midrule
\textit{Sex}   &                    &                     \\
Male           & 0.756              & 0.732               \\
Female         & 0.741              & 0.735               \\
\hline
\textbf{Gap}   & 0.015              & -0.003              \\ 
\midrule
\textit{BMI}   &                    &                     \\
Non-obese      & 0.752              & 0.746               \\
Obese          & 0.755              & 0.700               \\
\hline
\textbf{Gap}   & -0.003             & 0.046               \\ 
\bottomrule
\end{tabular}
\end{table}

\clearpage

\section{More experimental results} \label{sec:more_experiments}
\subsection{Ablation study}
\textbf{Impact of our core components} Table~\ref{tab:ablation_study} summarizes the sleep staging performance improvement of our principal components. This result demonstrates our individual components progressively improves the model's predictive power. The performance is evaluated on the SHHS1 test split. Experiments \edit{1-4} is pretrained on SHHS1 train split and experiments \edit{5-6} are done using the entire pretraining datasets. 

\textbf{Impact of demographic factors used in DGCL} Table~\ref{tab:demographic_ablation} presents the sleep staging accuracy with varying demographic factors used for DGCL. DGCL is only done for the training split of SHHS1 and accuracy is measured on SHHS1 test split.

\edit{
\textbf{Ablations on cycle length in DGCL} Table~\ref{tab:cycle_len_ablation} presents the sleep staging accuracy across varying cycle lengths used for DGCL. DGCL is applied only to the training split of SHHS1, and accuracy is measured on the SHHS1 test split.
}

\edit{
\textbf{Ablations on Macro-Encoder architecture} Table~\ref{tab:macro_encoder_ablation} presents the sleep staging accuracy with different macro-encoder architecture (uni-directional Mamba and vanilla Transformer). The bi-directional Mamba outperforms the other two architectures. The bi-directional Mamba shows its memory efficiency by reducing the memory footprint to half of that of the Transformer.
}

\edit{
\textbf{Ablations on pretraining strategy of Macro-Encoder}  We evaluated two well-known pretraining strategies: SimCLR~\cite{chen2020simple} and DINO~\cite{caron2021emerging}. In SimCLR, two augmented views are created by randomly shifting each subject’s full-night data by up to 60 epochs and randomly zero-masking 0 to 5 modalities without overlap between the two views. These two views from the same subject constitute a positive pair while all other views are considered negative. In DINO, the teacher model is constructed as the exponential moving average of the student model with a momentum of 0.996. The same augmentations are used to create two views which are used for the distillation process between the teacher and the student model. We kept the model architecture (bi-Mamba) identical for this experiment.
}

\edit{
The evaluation results are summarized in Table~\ref{tab:macro_encoder_pretrain_ablation}. While the alternative pretraining strategies demonstrate that it is possible to learn sleep macro-structure to some extent, DGCL achieves the strongest performance. We acknowledge, however, that the current distance function is empirically designed, and exploring learnable or adaptive metrics remains a promising direction for future work.
}

\begin{table}[ht]
    \centering
    \caption{\textbf{Ablation Study of \proposal Components.} We evaluate the contribution of each module and training strategy to the final sleep staging accuracy on the SHHS1 dataset.}
    \label{tab:ablation_study}
    \resizebox{0.75\linewidth}{!}{%
    \begin{tabular}{@{}c|cccccc|c@{}}
        \toprule
        \textbf{Id} & \textbf{Private encoder} & \textbf{Shared encoder} & \textbf{MAE} & \textbf{CL} & \textbf{Large scale pretraining} & \textbf{DGCL} & \textbf{Acc. (\%)} \\ \midrule
        1 & \checkmark &            & \checkmark &            &            &            & 74.8 \\
        2 & \checkmark & \checkmark & \checkmark &            &            &            & 75.7 \\
        3 & \checkmark & \checkmark & \checkmark & \checkmark &            &            & 76.0 \\
        \edit{4} & \checkmark & \checkmark &  & \checkmark &            &            & \edit{67.7} \\
        5 & \checkmark & \checkmark & \checkmark & \checkmark & \checkmark &            & 79.8 \\
        6 & \checkmark & \checkmark & \checkmark & \checkmark & \checkmark & \checkmark & \textbf{81.9} \\ \bottomrule
    \end{tabular}
    }
\end{table}

\begin{table}[H]
    \centering
    \caption{\textbf{Ablation Study of Demographic Factors in DGCL.} We investigate the impact of different demographic supervisory signals—Age, Sex, and BMI—on the final sleep staging accuracy. DGCL is only done on SHHS1 training split.}
    \label{tab:demographic_ablation}
    \resizebox{0.25\linewidth}{!}{%
    \begin{tabular}{@{}c|cccc@{}}
        \toprule
        \textbf{Id} & \textbf{Age} & \textbf{Sex} & \textbf{BMI} & \textbf{Acc. (\%)} \\ \midrule
        1 &            &            &            & 79.8 \\
        2 & \checkmark &            &            & 80.6 \\
        3 &            & \checkmark &            & 80.2 \\
        4 &            &            & \checkmark & 80.1 \\
        5 & \checkmark & \checkmark & \checkmark & \textbf{80.7} \\ \bottomrule
    \end{tabular}
    }
\end{table}

\begin{table}[h]
    \centering
    \caption{\edit{\textbf{Ablations on cycle length used for DGCL.} Performance is measured by sleep staging accuracy on the SHHS1 dataset.}}
    \label{tab:cycle_len_ablation}
    \resizebox{0.25\linewidth}{!}{%
    \begin{tabular}{@{}cc@{}}
        \toprule
        \textbf{Cycle length} & \textbf{Acc. (\%)} \\ \midrule
        60 & 80.5 \\
        \textbf{180} & \textbf{80.7} \\
        360 & 80.4 \\
        1080 & 80.3 \\
        \bottomrule
    \end{tabular}
    }
\end{table}

\begin{table}[h]
    \centering
    \caption{\edit{\textbf{Ablations on Macro-Encoder architecture.} 
    We compare the proposed Bi-directional Mamba architecture against Unidirectional Mamba and Transformer alternatives. Performance is measured by sleep staging accuracy on the SHHS1 dataset, alongside peak memory consumption during training. The Bi-directional Mamba achieves the highest classification accuracy while maintaining a modest memory footprint. All models were pretrained on training split of SHHS1 and a consistent batch size of 32 are used during training. Peak memory was measured on NVIDIA H100 HBM3 GPU.}
    }
    \label{tab:macro_encoder_ablation}
    \resizebox{0.6\linewidth}{!}{%
    \begin{tabular}{@{}ccc@{}}
        \toprule
        \textbf{Design choice} & \textbf{Acc. (\%)} & \textbf{Training peak memory (GB)} \\ \midrule
        Uni-directional Mamba & 78.7 & \textbf{1.84} \\
        \textbf{Bi-directional Mamba} & \textbf{79.7} & 4.01 \\
        Transformer & 79.2 & 7.78 \\        
        \bottomrule
    \end{tabular}
    }
\end{table}

\begin{table}[h]
    \centering
    \caption{\edit{\textbf{Ablations on pretrainig strategy of Macro-Encoder.}  DGCL demonstrates superior performance to other pretraining strategies (SimCLR~\cite{chen2020simple} and DINO~\cite{caron2021emerging}). Performance is measured by sleep staging accuracy and Macro-F1 on the SHHS1 dataset. All pretraining were done on the entire pretraining set.}}
    \label{tab:macro_encoder_pretrain_ablation}
    \resizebox{0.4\linewidth}{!}{%
    \begin{tabular}{@{}ccc@{}}
        \toprule
        \textbf{Pretraining Strategy} & \textbf{Acc. (\%)} &\textbf{Macro-F1}\\ \midrule
        \textbf{DGCL} & \textbf{81.9} & \textbf{74.1} \\
        SimCLR & 81.8 & 73.8 \\
        DINO & 81.7 & 73.7\\
        
        \bottomrule
    \end{tabular}
    }
\end{table}

\subsection{\edit{Alternative evaluation protocol - LSTM classifier}}
\edit{
While the evaluation results under the linear probing protocol are presented in Tables~\ref{tab:sleep_staging} and~\ref{tab:disease_results}, we also evaluate \proposal using an alternative protocol that utilizes an LSTM classifier, following SleepFM-Disease~\cite{thapa2026multimodal}. These results are summarized in Tables~\ref{tab:sleep_staging_lstm} and~\ref{tab:disease_prediction_lstm}. Notably, \proposal consistently outperforms all baselines under this alternative setup as well.
}

\subsection{\edit{Comparison to task-specialized models}}
\edit{
To evaluate the gap between foundation models and task-specialized approaches, we compare \proposal against sleep staging models (Table~\ref{tab:sleep_staging_lstm}) and an SDB segmentation specific model (Table~\ref{tab:sdb_segmentation_special}). We select SleePyCo~\cite{lee2024sleepyco}, SleepTransformer~\cite{phan2022sleeptransformer}, and DistillSleep~\cite{park2025distillsleep} for sleep staging, and AIX~\cite{hu2025transparent} for SDB segmentation, adapting the latter to support second-scale resolution. \proposal achieves highly competitive results. This strong performance is particularly notable considering that the specialized baselines undergo fully supervised training across all parameters, whereas \proposal relies on frozen encoder.
}

\begin{table}[h]
    \centering
    \caption{\edit{\textbf{Sleep stage classification results under alternative evaluation protocol and comparison to sleep staging specialized models.}  Following SleepFM, we use LSTM classifier during finetuning phase. In addition, sleep staging performance of three sleep staging models is provided.}}
    
    \label{tab:sleep_staging_lstm}
    \resizebox{0.6\linewidth}{!}{%
    \begin{tabular}{cccccc}
        \toprule
        \textbf{Dataset} & \textbf{Category} & \textbf{Models} & \textbf{Accuracy} & \textbf{Macro-F1} & \textbf{Kappa} \\ 
        \midrule
        \multirow{6}{*}{SHHS1} & Time series Foundation Model & MOMENT-Base & \textbf{85.2} & 77.4 & 78.8 \\
         \cmidrule(lr){2-6}
         & \multirow{2}{*}{\begin{tabular}[c]{@{}c@{}}Sleep\\ Foundation model\end{tabular}} & SleepFM-Disease & 83.8  & 78.0 & 77.4 \\
         
         &  & \textbf{\proposal} (Ours) & \textbf{85.2} & \textbf{79.0} & \textbf{79.1} \\ 
         \cmidrule(lr){2-6}
         & \multirow{3}{*}{\begin{tabular}[c]{@{}c@{}}Sleep Staging \\ Specialized Model\end{tabular}} & SleePyCo & 87.9 & 80.7 & 83.0 \\ 
         & & SleepTransformer & 87.7 & 80.1 & 82.8 \\
         & & DistillSleep-T & 86.8 & 81.1 & 81.4 \\
         
        \midrule
        \multirow{5}{*}{KISS} & Time series Foundation Model & MOMENT-Base & 77.4 & 76.5 & 70.5 \\
          \cmidrule(lr){2-6}
         & \multirow{2}{*}{\begin{tabular}[c]{@{}c@{}}Sleep\\ Foundation model\end{tabular}} & SleepFM-Disease & 74.6 & 75.8 & 67.8 \\
         &  & \textbf{\proposal} (Ours) & \textbf{77.7} & \textbf{77.3} & \textbf{71.0} \\ 
         \cmidrule(lr){2-6}
         & \multirow{2}{*}{\begin{tabular}[c]{@{}c@{}}Sleep Staging \\ Specialized Model\end{tabular}} & SleepTransformer & 77.8 & 77.2 & 71.1 \\
         & & DistillSleep-T & 80.3 & 80.0 & 74.5 \\
        \bottomrule
    \end{tabular}
    }
    \vspace{-1ex}
\end{table}

\begin{table}[h]
\centering
\caption{\edit{\textbf{Disease prediction results under the alternative evaluation protocol.} Following SleepFM, we use LSTM classifier during finetuning phase (SHHS1, C-Index).}}
\label{tab:disease_prediction_lstm}
\begin{tabular}{lcccccc}
\hline
\toprule
\textbf{Models} & \textbf{Angina} & \textbf{CVD death} & \textbf{CHF} & \textbf{CHD death} & \textbf{MI} & \textbf{Stroke} \\ 
\midrule
SleepFM         & 0.698           & 0.876              & 0.845        & 0.870              & 0.750       & 0.805           \\
\textbf{\proposal} & \textbf{0.702}  & \textbf{0.883}     & \textbf{0.854} & \textbf{0.891}     & \textbf{0.759} & \textbf{0.815}  \\ 
\bottomrule
\end{tabular}
\end{table}

\begin{table}[h]
    \centering
    \caption{\edit{\textbf{Comparison to SDB segmentation specialized model.}}}
    
    \label{tab:sdb_segmentation_special}
    \resizebox{0.4\linewidth}{!}{%
    \begin{tabular}{cccc}
        \toprule
        \textbf{Dataset} & \textbf{Models} & \textbf{Accuracy} & \textbf{Macro-F1}  \\ 
        \midrule
        \multirow{2}{*}{SHHS1} & AIX & \textbf{93.8} & \textbf{74.6} \\
         & \proposal (Ours) & 77.3 & 60.6 \\         
        \bottomrule
    \end{tabular}
    }
    \vspace{-1ex}
\end{table}

\subsection{More embedding analysis} \label{sec:more_viz}
\begin{figure}[H]
    \centering
    \begin{subfigure}[b]{0.45\columnwidth}
        \centering
        \includegraphics[width=0.99\textwidth, bb=0 0 209 220]{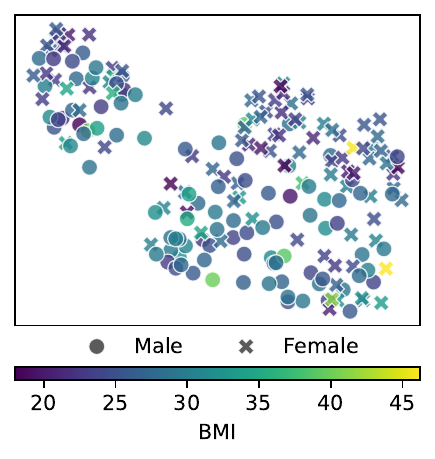}
        \caption{SHHS1 - BMI and Sex}
        \label{fig:umap_shhs_bmisex}
    \end{subfigure}    
    \begin{subfigure}[b]{0.45\columnwidth}
        \centering
        \includegraphics[width=0.99\textwidth, bb=0 0 209 220]{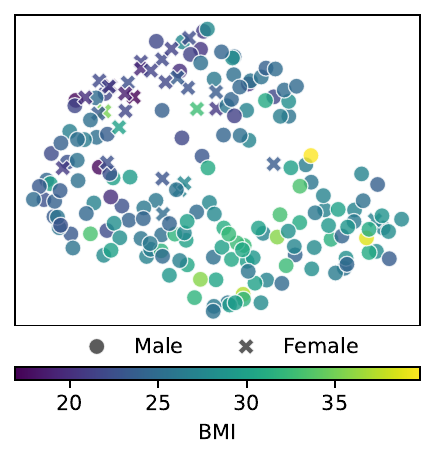}
        \caption{KISS - BMI and Sex}
        \label{fig:umap_kiss_bmisex} 
    \end{subfigure}
    \begin{subfigure}[b]{0.45\columnwidth}
        \centering
        \includegraphics[width=0.99\textwidth, bb=0 0 209 220]{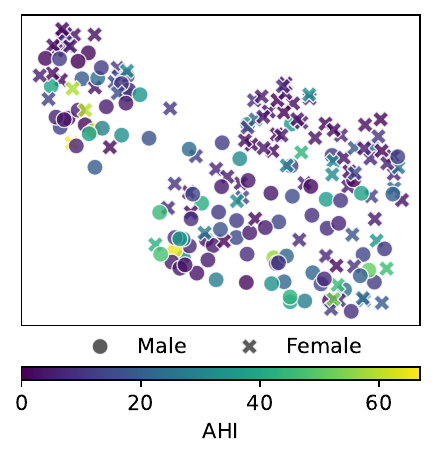}
        \caption{SHHS1 - AHI and Sex}
        \label{fig:umap_shhs_ahiisex}
    \end{subfigure}    
    \begin{subfigure}[b]{0.45\columnwidth}
        \centering
        \includegraphics[width=0.99\textwidth, bb=0 0 209 220]{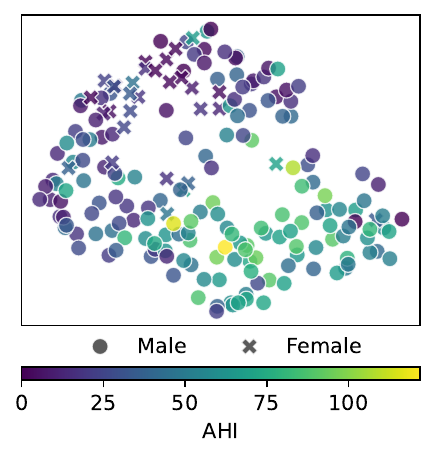}
        \caption{KISS - AHI and Sex}
        \label{fig:umap_kiss_ahisex} 
    \end{subfigure}
    \caption{\textbf{Macro-embeddings visualization.} Per-subject latent embeddings from Macro-Encoder is visualized using U-MAP. Each point represents a single subject's embeddings where color and symbol represents demographic characteristics.}
    \label{fig:umap_appendix}
    \vspace{-2ex}
\end{figure}

Figure~\ref{fig:umap_appendix} presents more visualizations of Macro-embeddings similar to those presented in Figure~\ref{fig:umap} using different demographic features.

\clearpage

\clearpage

\subsection{Confusion matrix}
\begin{figure}[H]
    \centering
    \begin{subfigure}[b]{0.45\columnwidth}
        \centering
        \includegraphics[width=0.99\textwidth, bb=0 0 323 323]{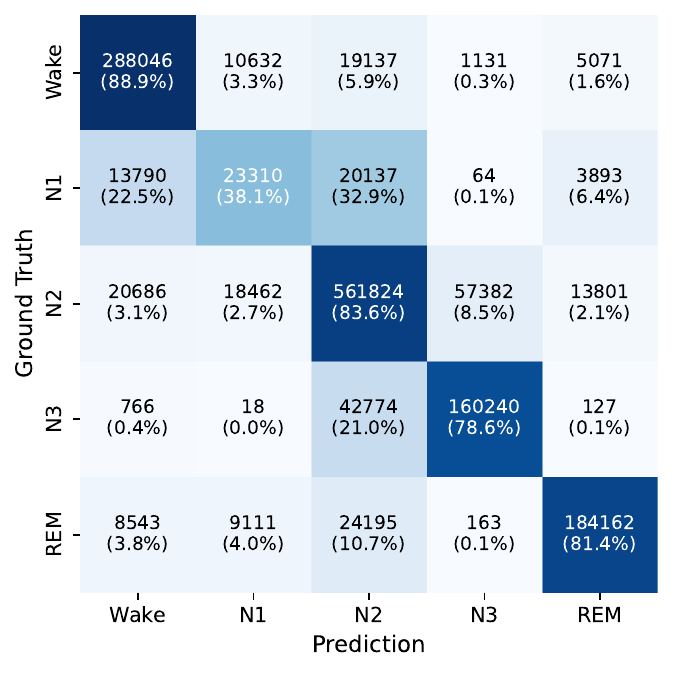}
        \caption{SHHS1 - sleep staging}
        \label{fig:conf_shhs1}
    \end{subfigure}    
    \begin{subfigure}[b]{0.45\columnwidth}
        \centering
        \includegraphics[width=0.99\textwidth, bb=0 0 323 323]{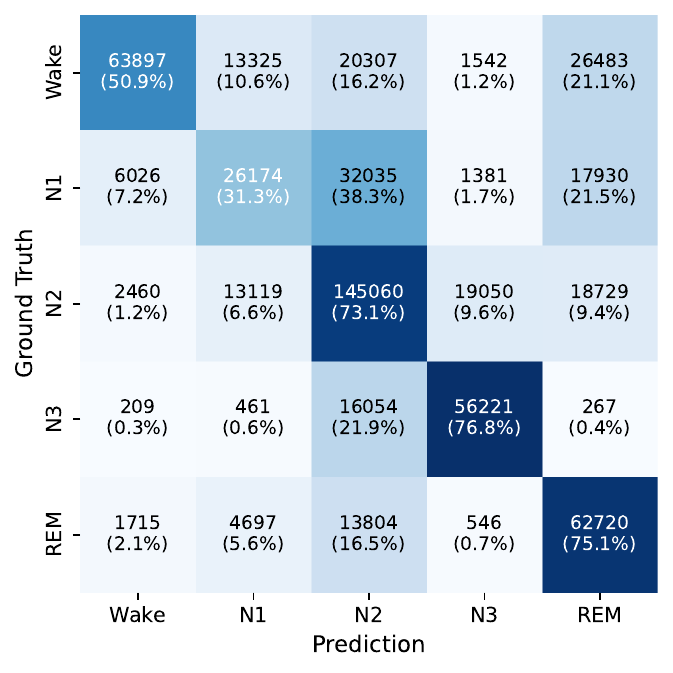}
        \caption{KISS - sleep staging}
        \label{fig:conf_kiss} 
    \end{subfigure}
    \begin{subfigure}[b]{0.45\columnwidth}
        \centering
        \includegraphics[width=0.99\textwidth, bb=0 0 323 323]{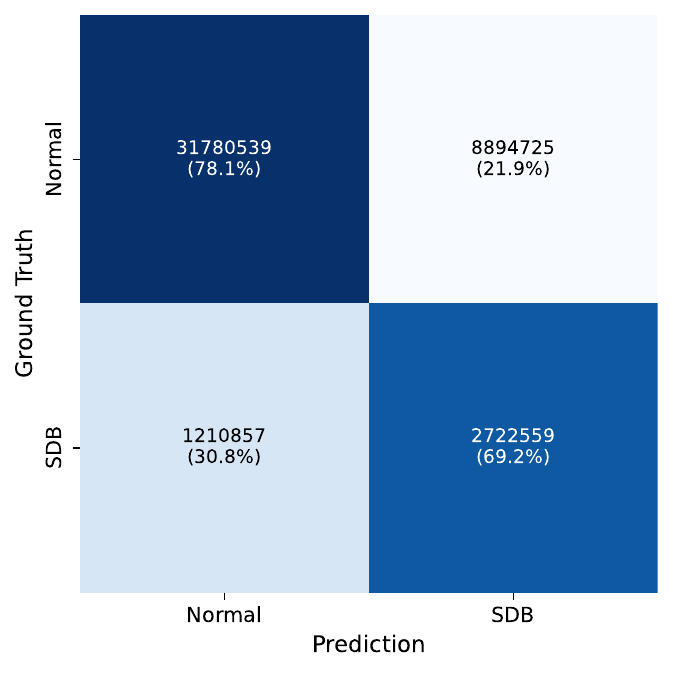}
        \caption{SHHS1 - SDB segmentation}
        \label{fig:conf_shhs1_ah}
    \end{subfigure}    
    \begin{subfigure}[b]{0.45\columnwidth}
        \centering
        \includegraphics[width=0.99\textwidth, bb=0 0 323 323]{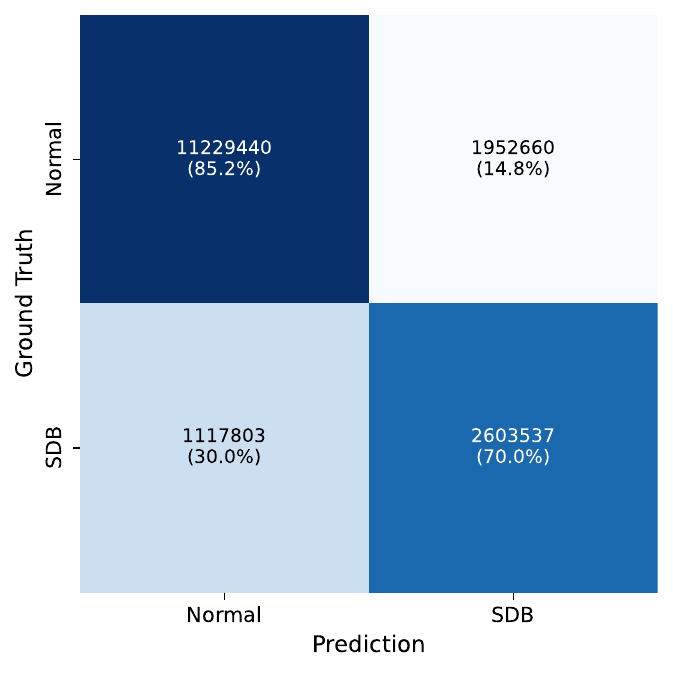}
        \caption{KISS - SDB segmentation}
        \label{fig:conf_kiss_ah} 
    \end{subfigure}
    \caption{Confusion matrix for sleep staging and SDB segmentation on the test split of SHHS1 and KISS.}
    \label{fig:conf}
    \vspace{-2ex}
\end{figure}

In Figure~\ref{fig:conf}, we provide the confusion matrix for sleep staging and SDB segmentation evaluated on the test split of SHHS1 and KISS.

\end{document}